\documentclass[10pt,journal,twoside]{IEEEtran}
\usepackage{amsmath,amsfonts}
\usepackage{algorithmic}
\usepackage{array}
\usepackage[caption=false,font=scriptsize,labelfont=sf,textfont=sf]{subfig}
\usepackage{textcomp}
\usepackage{stfloats}
\usepackage{url}
\usepackage{verbatim}
\usepackage{graphicx}
\usepackage{cite}
\usepackage[numbers]{natbib}
\usepackage{booktabs}
\usepackage{multirow}
\usepackage{makecell}
\usepackage{tabularx}
\usepackage{tablefootnote}
\usepackage{threeparttable}
\usepackage{caption}
\usepackage{pifont}
\usepackage{tikz}
\usepackage[T1]{fontenc} 

\newcommand{\Rmnum}[1]{\uppercase\expandafter{\romannumeral#1}}
\newcommand{\indep}{\perp\!\!\!\!\!\perp}
\newcommand{\parahead}[1]{\vskip 0.6em\noindent\textbf{#1}~~}
\newcommand{\paraheadtop}[1]{\noindent\textbf{#1}~~}

\definecolor{myblue}{HTML}{95d0fc}    
\definecolor{myyellow}{HTML}{e6daa6}  
\definecolor{mygreen}{HTML}{087804}   
\definecolor{myred}{HTML}{980002}     

\begin{document}

\title{Revisiting Label Inference Attacks in Vertical Federated Learning: Why They Are Vulnerable and How to Defend}

\author{
    Yige Liu, Dexuan Xu, Zimai Guo, Yongzhi Cao, \textit{Senior Member, IEEE}, and Hanpin Wang
    \thanks{Yige Liu, Dexuan Xu, Zimai Guo, Yongzhi Cao, and Hanpin Wang are with the Key Laboratory of High Confidence Software Technologies (Peking University), Ministry of Education; School of Computer Science, Peking University, Beijing, China.~(e-mail: yige.liu@stu.pku.edu.cn; xudexuan@stu.pku.edu.cn; endergzm@pku.edu.cn; caoyz@pku.edu.cn; whpxhy@pku.edu.cn)}
    \thanks{Yongzhi Cao is also with the Zhongguancun Laboratory, Beijing, China.}
    \thanks{Corresponding author: Yongzhi Cao}
}

\markboth{}{}

\maketitle

\begin{abstract}
    Vertical federated learning~(VFL) allows an active party with a top model, and multiple passive parties with bottom models to collaborate. In this scenario, passive parties possessing only features may attempt to infer active party's private labels, making label inference attacks~(LIAs) a significant threat. Previous LIA studies have claimed that well-trained bottom models can effectively represent labels. However, we demonstrate that this view is misleading and exposes the vulnerability of existing LIAs. By leveraging mutual information, we present the first observation of the ``model compensation'' phenomenon in VFL. We theoretically prove that, in VFL, the mutual information between layer outputs and labels increases with layer depth, indicating that bottom models primarily extract feature information while the top model handles label mapping. Building on this insight, we introduce task reassignment to show that the success of existing LIAs actually stems from the distribution alignment between features and labels. When this alignment is disrupted, the performance of LIAs declines sharply or even fails entirely. Furthermore, the implications of this insight for defenses are also investigated. We propose a zero-overhead defense technique based on layer adjustment. Extensive experiments across five datasets and five representative model architectures indicate that shifting cut layers forward to increase the proportion of top model layers in the entire model not only improves resistance to LIAs but also enhances other defenses.
\end{abstract}


\section{Introduction}

The growing challenge of data scarcity and the increasing demand for privacy preservation have boosted the federated learning, a distributed machine learning paradigm that enables multiple participants to collaboratively train a model without sharing their data~\cite{mcmahan2016federated,wen2023survey}. Based on the division of participants' data~\cite{yang2019federated}, federated learning is categorized into horizontal federated learning~\cite{ayeelyan2024federated,huang2024federated,yazdinejad2024robust}, vertical federated learning~(VFL)~\cite{liu2023vertical,liu2024label,ye2025vertical}, and federated transfer learning~\cite{guo2024comprehensive,ma2024fedst,zhao2025personalized}. In VFL, data is distributed among multiple passive parties and one active party. The passive parties share the same sample space but possess different sample features without any labels, while the active party holds only the sample labels. During training, passive parties use their bottom models to transform features into embeddings and send these embeddings to the active party. Upon receiving the embeddings from all passive parties, the active party aggregates them and predicts their labels using the top model. This process enables the active party to complete the supervised training of the VFL model based on its labels.

Therefore, in such scenarios, passive parties possessing only features may attempt to infer the private labels held by the active party. This type of attack is known as the label inference attack~(LIA)~\cite{wainakh2021label,kariyappa2023exploit,zhang2022data,liu2024similarity,fu2022label}. When launching LIAs, honest-but-curious attackers seek to exploit any available information to facilitate their attack. A widely adopted and effective strategy is to infer labels from embeddings generated by well-trained bottom models. This approach relies on an intuitive assumption~(which we term the embedding-label assumption): well-trained bottom models establish a strong mapping between their generated embeddings and the labels. As a result, inferring label categories by the clustering properties of embeddings or leveraging the bottom model's label representation capability forms the foundation of many LIAs~\cite{sun2022label,liu2024similarity,fu2022label}.

However, in this paper, we find that this intuitive assumption may mislead existing LIA research. By leveraging mutual information, we first quantify the correlation between the outputs of each layer and the labels in VFL. We observe that the mutual information between each layer's outputs and labels in the bottom model is significantly lower than that in the top model, indicating that the bottom model has poor label representation capability. To support this observation, we provide a theoretical proof that the mutual information between each layer's outputs and labels increases with layer depth in VFL. This finding clarifies the actual roles of the bottom and top models, where the bottom models primarily focus on feature extraction, while the top model is responsible for label mapping. Furthermore, we also present the first observation of the ``model compensation'' phenomenon on the top model in VFL. Specifically, as the number of passive parties increases, each passive party holds relatively fewer features. This further weakens the label representation capability of each bottom model, forcing the top model to compensate for the label mapping function. In other words, the multi-party nature of VFL amplifies the functional divergence between the bottom and top models, causing the bottom models to become increasingly specialized in feature extraction, while the top model assumes a more critical role in label mapping.

Based on our analysis of the actual roles of the bottom and top models, we contend that the apparent success of existing LIAs grounded in the embedding-label assumption may be largely coincidental. To examine this, we design task reassignment experiments that deliberately disrupt the latent mapping between sample features and their original labels. Experimental results on LIAs across three new tasks demonstrate that the high attack accuracy of existing LIA methods is largely an illusion, heavily dependent on a strong natural alignment between sample features and labels. When this alignment is disrupted, model compensation in VFL paradoxically amplifies the bottom model's focus on feature representation rather than label mapping, undermining the foundation of existing LIAs. This exposes the vulnerability of these LIAs.

Inspired by the above findings, we further explore and propose a novel defense technique: advancing the cut layer that connects the bottom and top models to increase the proportion of the top model in the overall architecture. This adjustment weakens the association between the bottom model and the labels, thereby providing effective defense against LIAs. We conduct experiments across five distinct model architectures on three benchmark datasets and two real-world datasets. The results show that advancing the cut layer by just one layer yields substantial defense benefits, while advancing it by three layers reduces LIA attack accuracy to the level of random guessing. Moreover, we observe that this defense technique not only provides the robust privacy preservation but also enhances the predictive performance of VFL models. Experiments with three general defense strategies and two specific defenses further demonstrate that this technique can significantly enhance the effectiveness of existing defense strategies.

Our main contributions are summarized as follows:
\begin{itemize}
    \item We theoretically demonstrate that the mutual information between each layer's outputs and the labels increases with layer depth, which further clarifies the actual roles of the bottom and top models. We also present the first observation of the ``model compensation'' phenomenon in VFL and find that it further promotes the functional differentiation between bottom and top models.
    \item Through task reassignment, we revisit existing LIAs and reveal that their vulnerabilities arise from an overreliance on the natural alignment between sample features and labels, which may inspire future research.
    \item We propose a zero-overhead defense technique that defends against LIAs by moving forward the cut layers, which can enhance model prediction performance while also strengthening other defense strategies.
\end{itemize}


\section{Preliminaries}

\subsection{Vertical Federated Learning}

\begin{figure}[t!]
    \centering
    \includegraphics[width=\columnwidth]{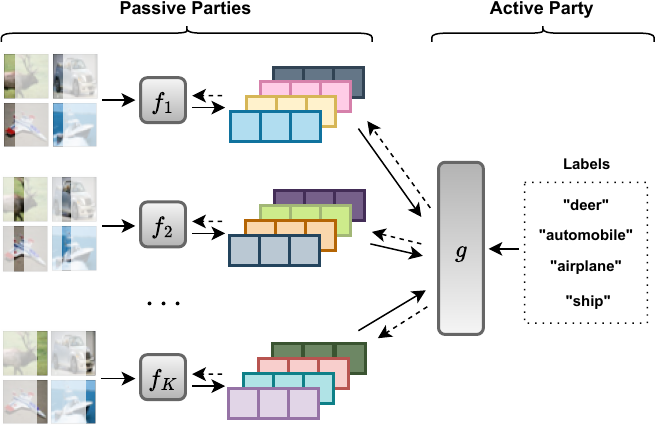}
    \caption{An illustration of the vertical federated learning framework. Solid lines represent forward propagation and dashed lines denote backpropagation.}
    \label{fig:vfl}
\end{figure}

VFL is a distributed machine learning paradigm that allows $K$ passive parties holding different features and an active party holding labels to collaboratively train a model with $N$ samples. Typically, the last layer of the bottom model and the first layer of the top model are referred to as the cut layers~\cite{vepakomma2018split}. Based on whether the active party possesses additional features, VFL can be categorized into two types: aggVFL and splitVFL~\cite{liu2023vertical}. In aggVFL, the active party holds only labels, while in splitVFL, it holds both labels and features. As illustrated in Fig.~\ref{fig:vfl}, we concentrate on aggVFL in the paper and formalize its training process as follows.

During forward propagation, each passive party $k$ uses its bottom model $f_k\left(\cdot\right)$ parameterized by $\theta_k$ and its local data $x_i^k$ from datasets $\left\{x_i^k\right\}_{i=1}^N$ to calculate embeddings $z_i^k=f_k\left(\theta_k;x_i^k\right)$. These embeddings are then sent to the active party for training. Upon receiving the embeddings from all passive parties, the active party first aggregates these embeddings into a single embedding $z_i=\mathrm{agg}(z_i^1,\dots,z_i^K)$ by a certain aggregation algorithm. Subsequently, it trains the top model $g\left(\cdot\right)$ parameterized by $\varphi$ using the aggregated embeddings $\{z_i\}_{i=1}^N$ and ground-truth labels $\{y_i\}_{i=1}^N$ with the loss function
\begin{equation}
    \mathcal{L}\left(\theta,\varphi;x,y\right)=\frac{1}{N} \sum_{i=1}^N \ell\left(g\left(\varphi;z_i\right),y_i\right),
\end{equation}
where $\ell\left(\cdot\right)$ denotes the sample loss such as cross-entropy loss or mean squared error loss.

During backpropagation, the active party calculates gradients $\nabla_{\varphi}\mathcal{L}$ to update the top model and sends gradients $\nabla_{z_i^k}\mathcal{L}=\frac{\partial\mathcal{L}}{\partial z_i^k}$ to each passive party. Leveraging the chain rule, each passive party then updates its bottom model with these gradients as
\begin{equation}
    \nabla_{\theta_k}\mathcal{L} = \frac{\partial\mathcal{L}}{\partial z_i^k} \cdot \frac{\partial z_i^k}{\partial\theta_k} = \nabla_{z_i^k}\mathcal{L} \cdot \frac{\partial z_i^k}{\partial\theta_k},
\end{equation}
\begin{equation}
    \theta_k^{t+1}=\theta_k^t - \eta\cdot\nabla_{\theta_k}\mathcal{L},
\end{equation}
where $t$ denotes the iteration and $\eta$ denotes the learning rate.

The entire training process is iterated until the model converges.

\subsection{Mutual Information}

The mutual information~(MI)~\cite{shannon1948mathematical} between two discrete random variables $X$ and $Y$ is defined to be
\begin{equation}
    I(X;Y)=\sum_{x\in X, y\in Y}p(x,y)\log\left(\frac{p(x,y)}{p(x)p(y)}\right),
\end{equation}
where $p(x,y)$ is the joint probability distribution function of $X$ and $Y$, and $p(x)$ and $p(y)$ are the marginal probability distribution functions of $X$ and $Y$, respectively. Intuitively, mutual information quantifies the amount of information obtained about one random variable through observing the other random variable. A smaller mutual information indicates a weaker correlation between two random variables, while a larger mutual information reflects a stronger correlation. In addition, it can also be expressed in terms of entropy as
\begin{equation}
    I(X;Y)=H(X)-H(X|Y)=H(Y)-H(Y|X),
\end{equation}
where $H(X)$ is the entropy of $X$ and $H(X|Y)$ is the conditional entropy of $X$ given $Y$.

\subsection{Label Inference Attacks}
\label{sec:lia}

LIA is specific to VFL scenarios, where the passive party lacks labels and attempts to steal the private labels holding by the active party through the training process~\cite{li2022label}. Due to the unique characteristics of VFL scenarios, successful label inference is often a prerequisite for launching many subsequent attacks. For example, in backdoor attacks~\cite{bai2023villain}, adversaries aiming to perform targeted backdoor operations must first infer the labels of samples before implanting the backdoor. As a result, LIA is considered both representative and critically important within the VFL.

Based on attack methods, \citet{liu2024label} classified LIAs into several categories. Among these, two embedding-based approaches---LIA with classification and cluster~\cite{sun2022label,liu2024similarity} and LIA with model completion~\cite{fu2022label}---have been extensively studied, as they are grounded in more practical rather than strict security assumptions. Both of these LIAs rely on the intuitive embedding-label assumption: that embeddings generated by a well-trained bottom model exhibit a strong mapping to the labels. In the paper, we focus on these LIAs, which primarily exploit the intuitive embedding-label assumption, and reveal their vulnerabilities. Utilizing this assumption, \citet{fu2022label} proposed an attack method that fine-tunes the bottom model with a small amount of labeled auxiliary data, enabling the fine-tuned bottom model to infer labels. \citet{sun2022label} leveraged this with a spectral attack~\cite{tran2018spectral} to distinguish the distribution of embeddings between different labels. In addition, \citet{bai2023villain} found that the top model returns a larger gradient when an embedding is replaced with one from a different label category. Thus, a gradient classifier can be used to infer label categories by replacing embeddings. Moreover, \citet{liu2024similarity} utilized the K-means algorithm and Euclidean distance to classify embeddings and infer labels, as embeddings corresponding to different labels form distinct categories.


\section{Actual Roles of Bottom and Top Models}
\label{sec:model_compensation}

In this section, we combine experimental results and theoretical analysis to clarify the practical roles of the bottom and top models in VFL: the bottom model focuses on feature extraction, while the top model handles the label mapping.

\subsection{Settings}
\label{sec:model_comp_setup}

\paraheadtop{Datasets.} We carefully select five datasets, including three benchmark datasets MNIST~\cite{lecun1998gradient}, CIFAR-10~\cite{krizhevsky2009learning}, and CINIC-10~\cite{darlow2018cinic10}, and two real-world datasets SVHN~\cite{netzer2011reading} and GTSRB~\cite{stallkamp2011german}. These datasets have been widely used in previous research on attacks and defenses in federated learning~\cite{krauss2023mesas,pang2025poisafl,fan2025boosting,xu2025risk}. Among them, SVHN contains street view house numbers, while GTSRB is the German traffic sign recognition dataset. Except for the MNIST dataset, which consists of grayscale images, all other datasets comprise color images. In addition, the first four datasets each contain 10 classes, whereas GTSRB includes 43 classes. We use random sampling along columns to divide features, simulating how passive parties in VFL share the sample space while retaining distinct feature spaces.

\parahead{Models.} The entire model follows the VFL framework illustrated in Fig.~\ref{fig:vfl}. On the passive party side, we adopt different bottom model architectures for each dataset: a multilayer perceptron~(MLP)~\cite{gardner1998artificial} for MNIST, AlexNet~\cite{krizhevsky2012imagenet} for CIFAR-10, ResNet~\cite{he2016deep} for CINIC-10, LeNet~\cite{lecun1998gradient} for SVHN, and a common convolutional neural network~(CNN) for GTSRB. On the active party side, the aggregation algorithm combines the embeddings uploaded by different passive parties according to their dimensions. For the top model, we utilize the multilayer perceptron for all datasets.

\begin{figure*}[t!]
    \centering
    \subfloat[MNIST, MLP]{
        \includegraphics[width=0.19\textwidth]{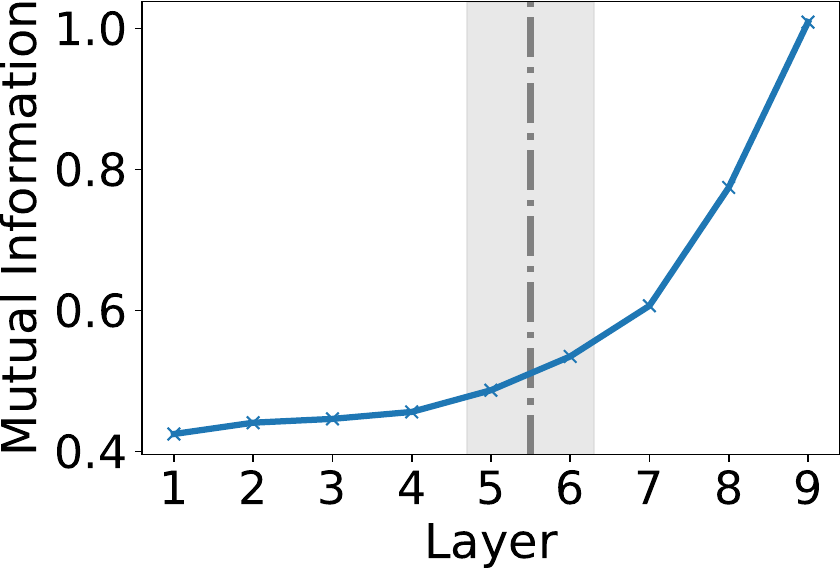}
        \hspace{0.01mm}
        \includegraphics[width=0.19\textwidth]{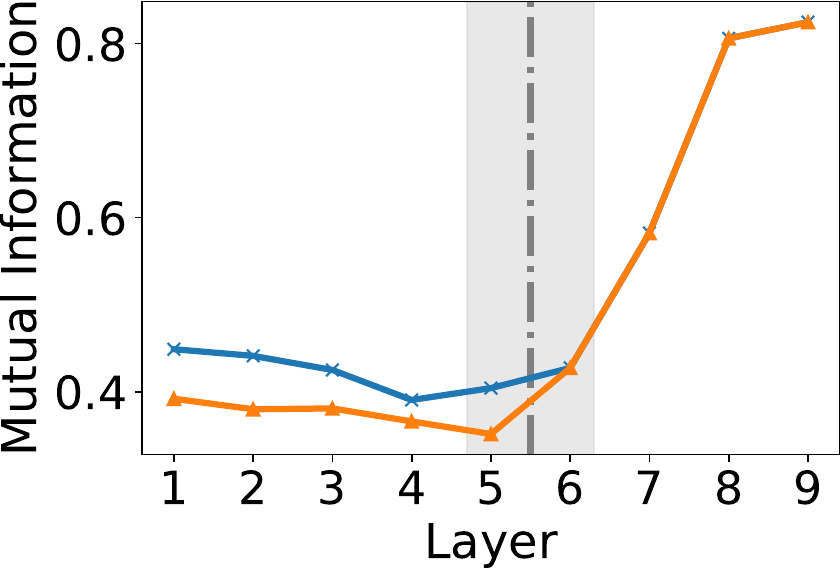}
        \hspace{0.01mm}
        \includegraphics[width=0.19\textwidth]{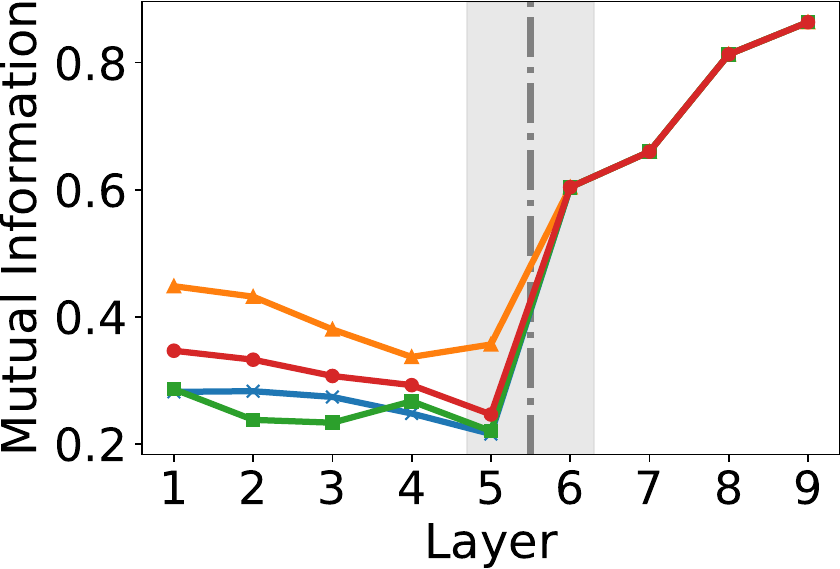}
        \hspace{0.01mm}
        \includegraphics[width=0.19\textwidth]{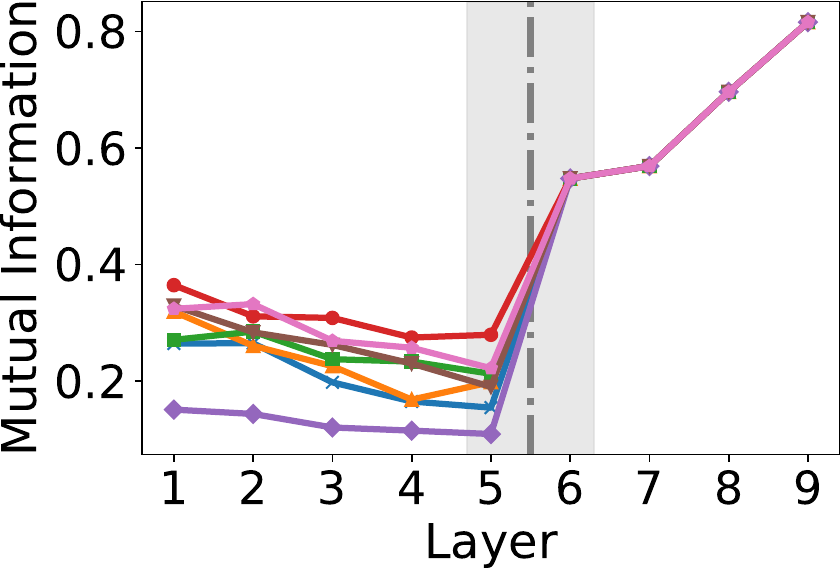}
        \hspace{0.01mm}
        \includegraphics[width=0.19\textwidth]{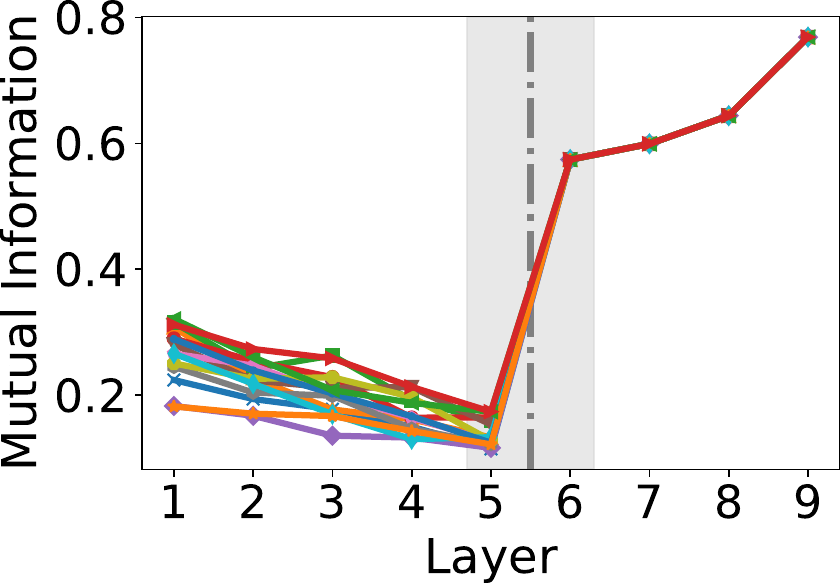}
    }\\
    \subfloat[CIFAR-10, AlexNet]{
        \includegraphics[width=0.19\textwidth]{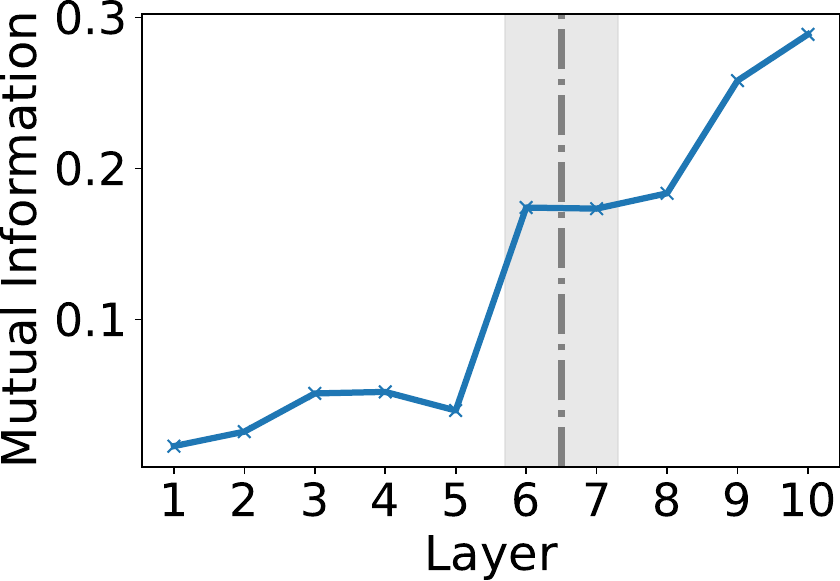}
        \hspace{0.01mm}
        \includegraphics[width=0.19\textwidth]{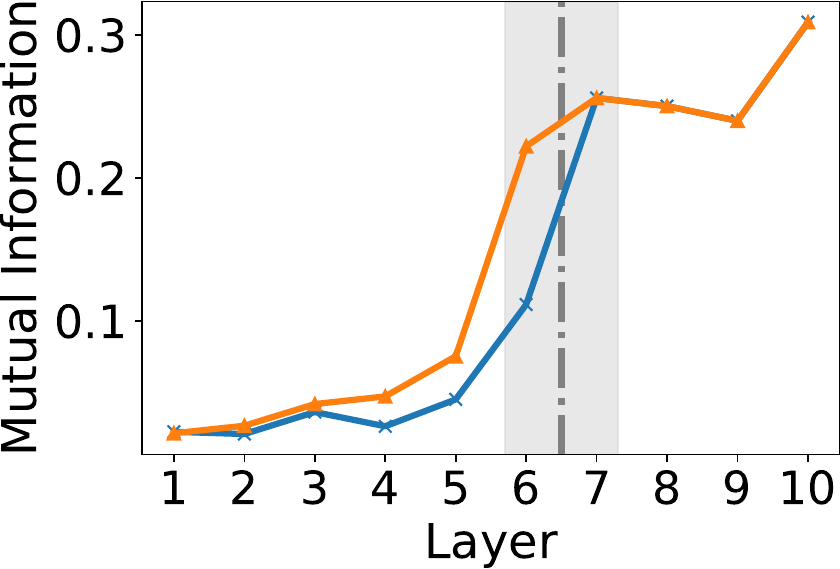}
        \hspace{0.01mm}
        \includegraphics[width=0.19\textwidth]{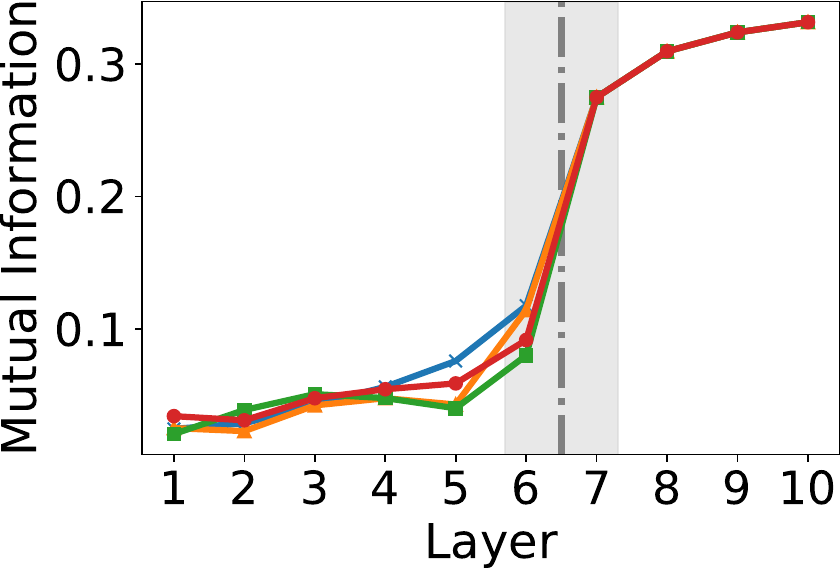}
        \hspace{0.01mm}
        \includegraphics[width=0.19\textwidth]{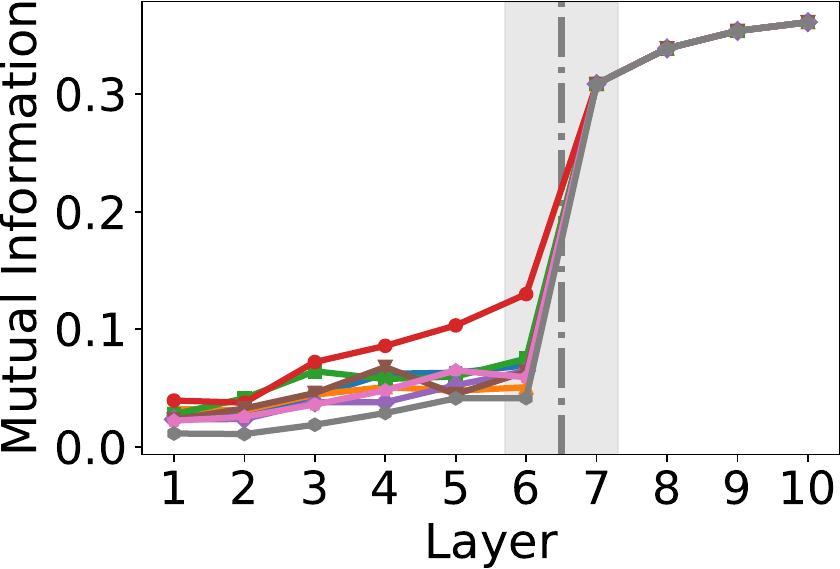}
        \hspace{0.01mm}
        \includegraphics[width=0.19\textwidth]{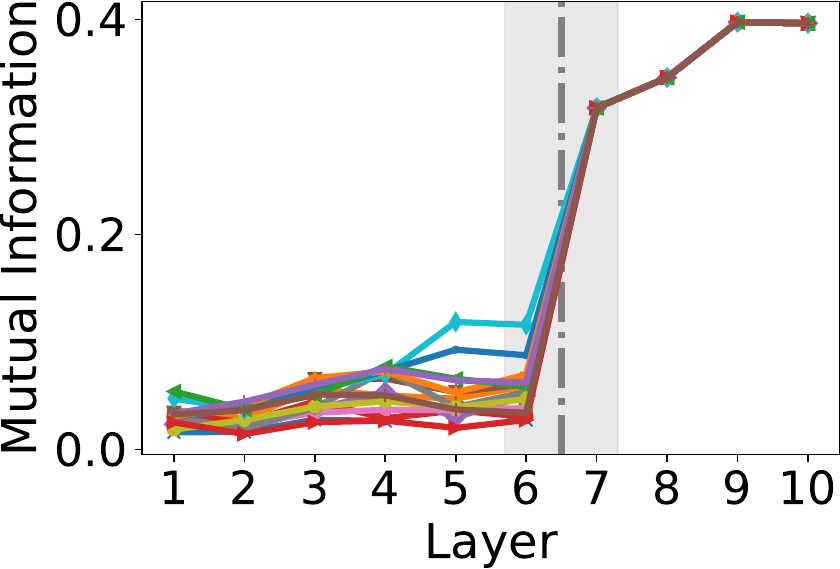}            
    }\\
    \subfloat[CINIC-10, ResNet]{
        \includegraphics[width=0.19\textwidth]{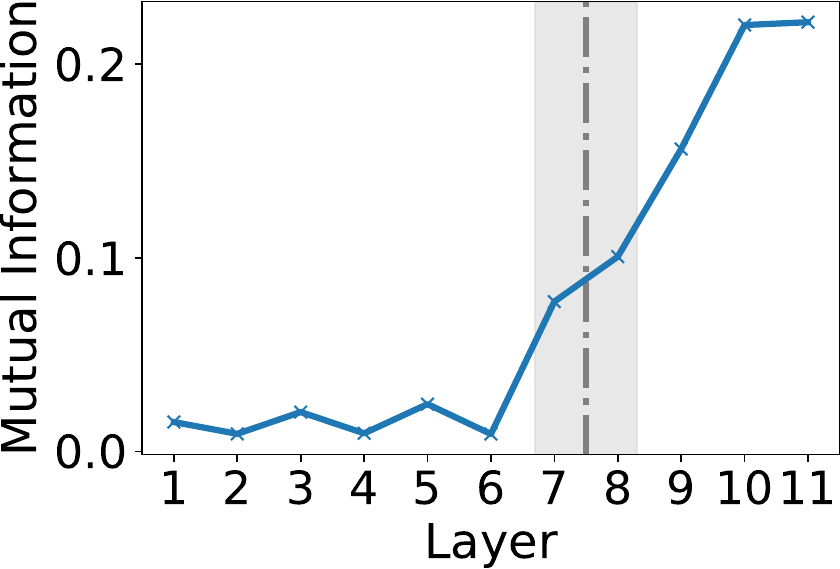}
        \hspace{0.01mm}
        \includegraphics[width=0.19\textwidth]{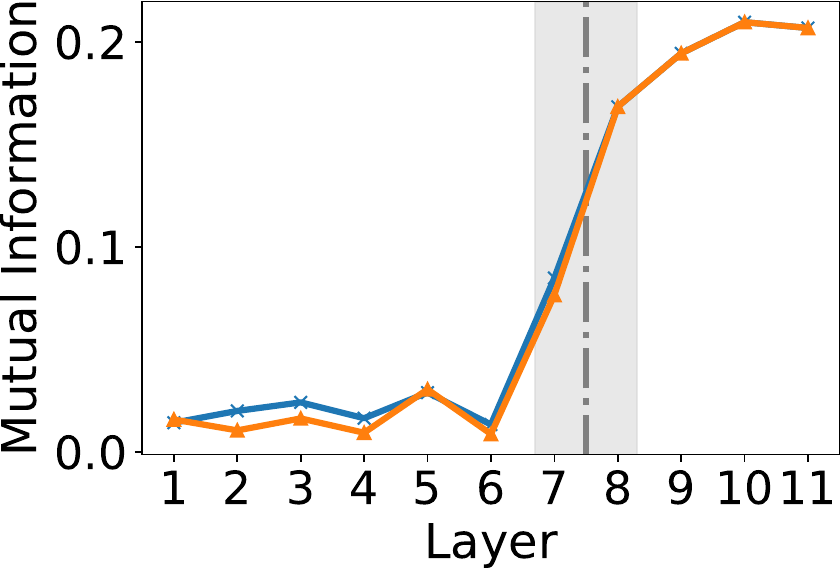}
        \hspace{0.01mm}
        \includegraphics[width=0.19\textwidth]{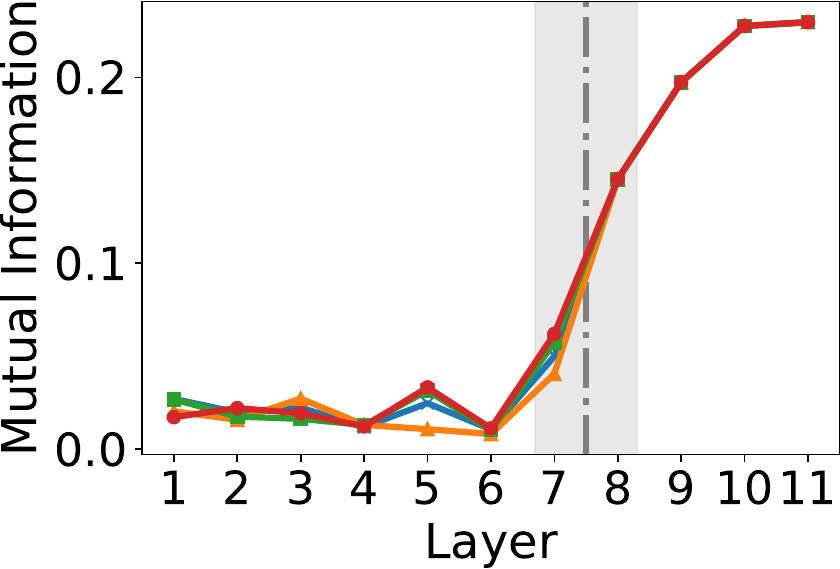}
        \hspace{0.01mm}
        \includegraphics[width=0.19\textwidth]{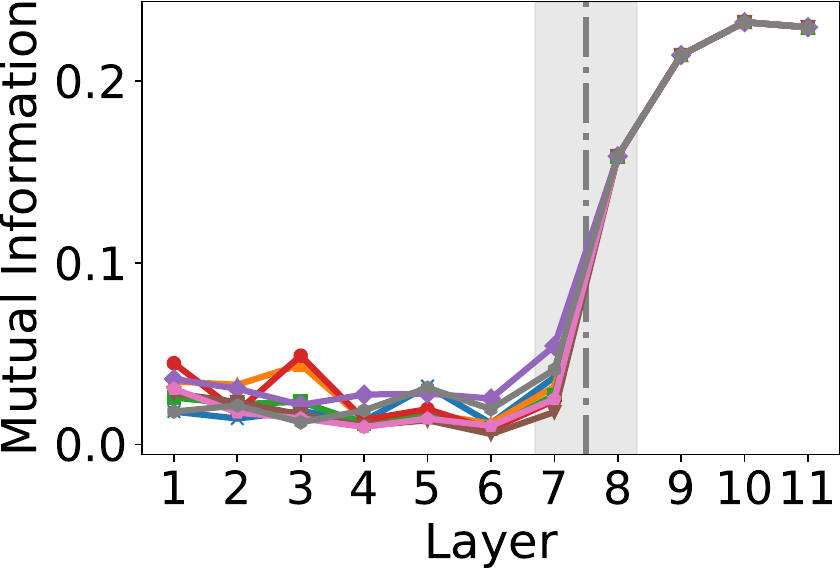}
        \hspace{0.01mm}
        \includegraphics[width=0.19\textwidth]{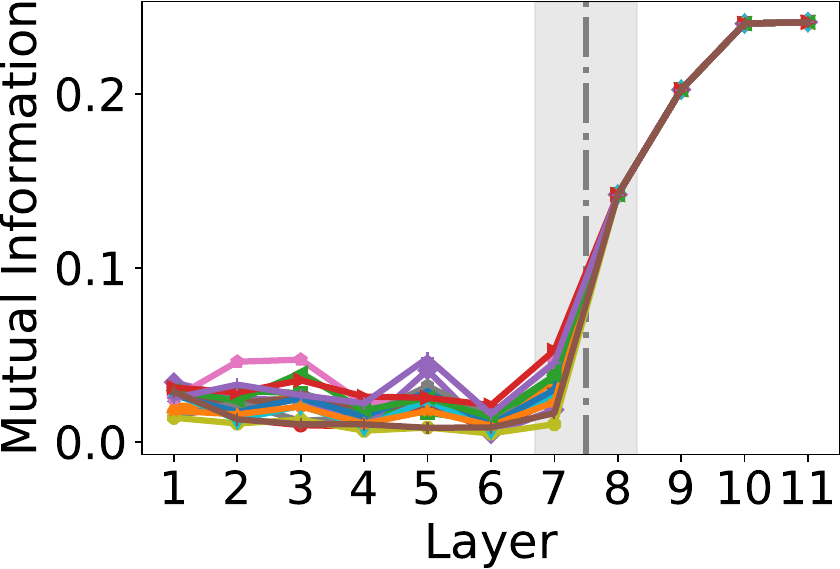}
    }\\
    \subfloat[SVHN, LeNet]{
        \includegraphics[width=0.19\textwidth]{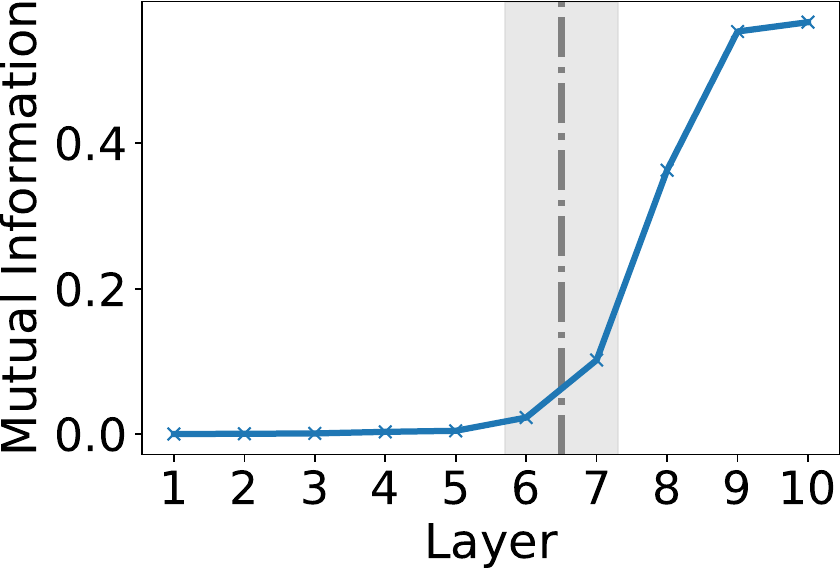}
        \hspace{0.01mm}
        \includegraphics[width=0.19\textwidth]{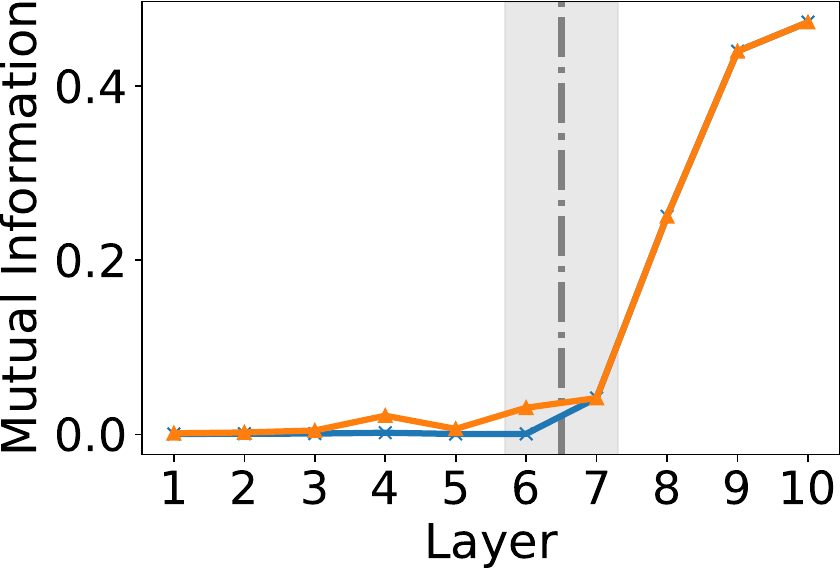}
        \hspace{0.01mm}
        \includegraphics[width=0.19\textwidth]{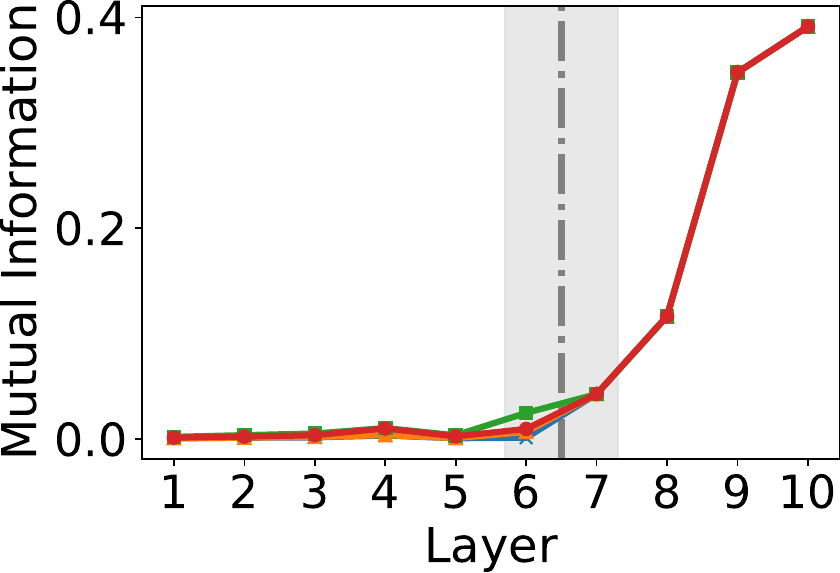}
        \hspace{0.01mm}
        \includegraphics[width=0.19\textwidth]{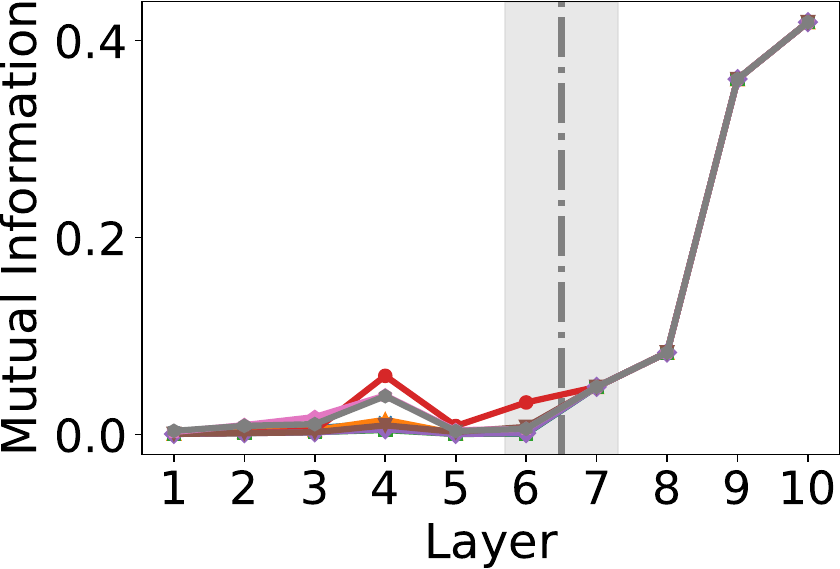}
        \hspace{0.01mm}
        \includegraphics[width=0.19\textwidth]{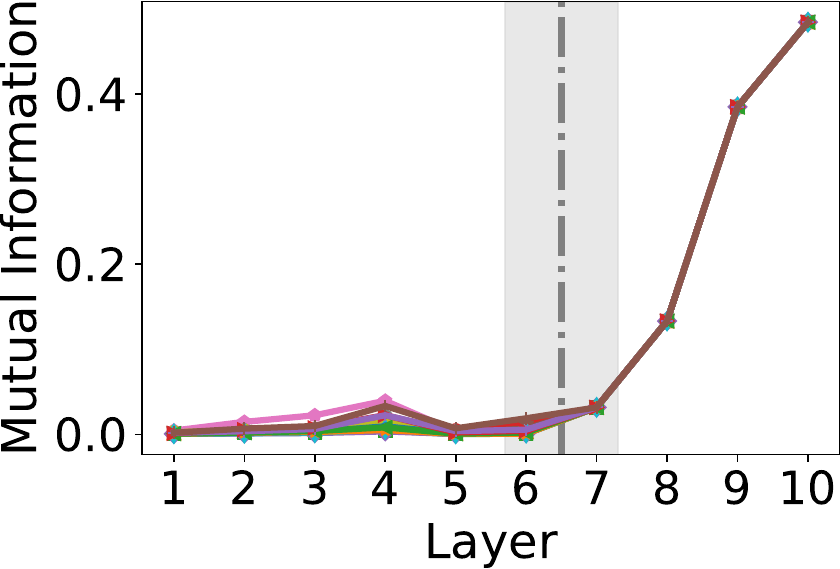}
    }\\
    \subfloat[GTSRB, CNN]{
        \includegraphics[width=0.19\textwidth]{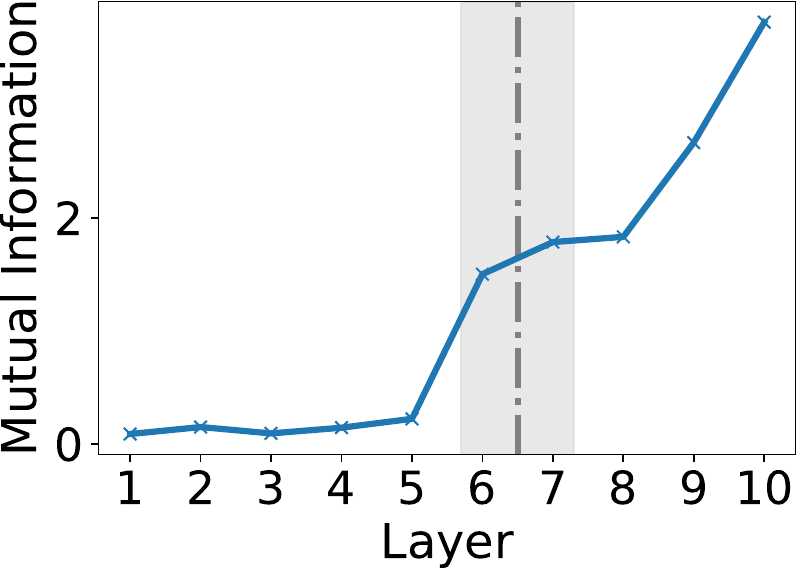}
        \hspace{0.01mm}
        \includegraphics[width=0.19\textwidth]{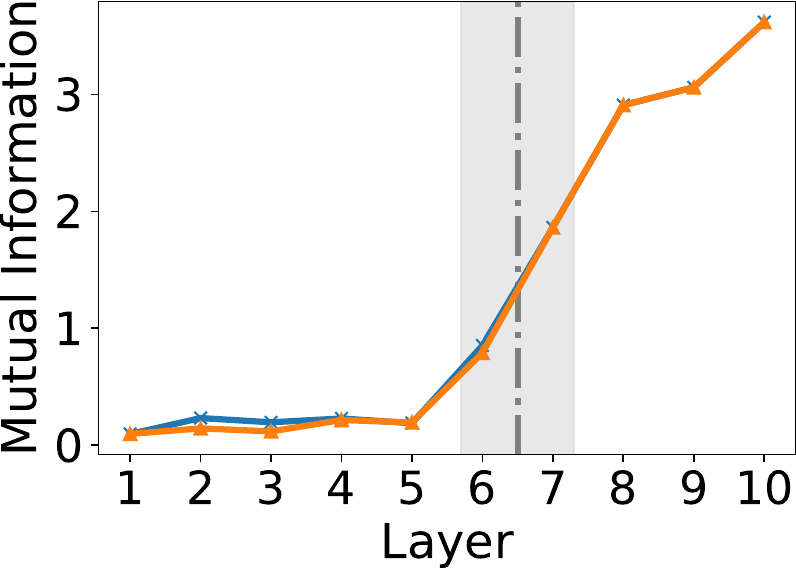}
        \hspace{0.01mm}
        \includegraphics[width=0.19\textwidth]{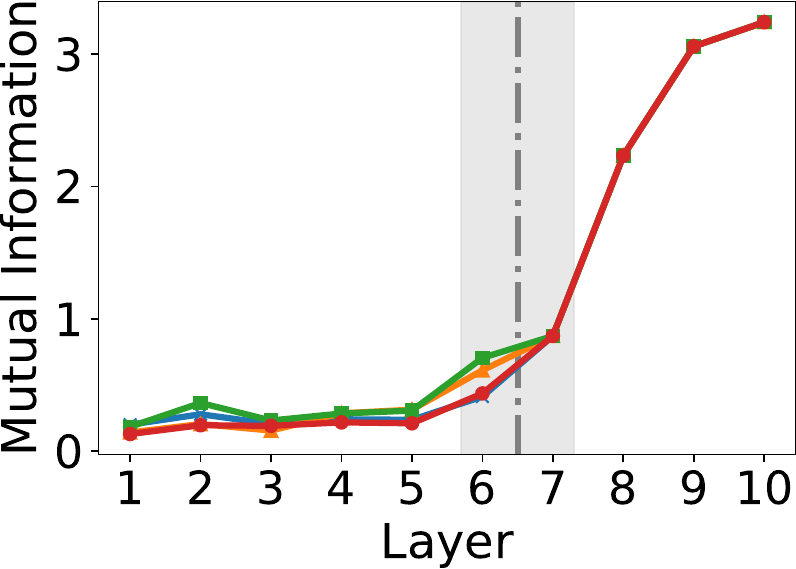}
        \hspace{0.01mm}
        \includegraphics[width=0.19\textwidth]{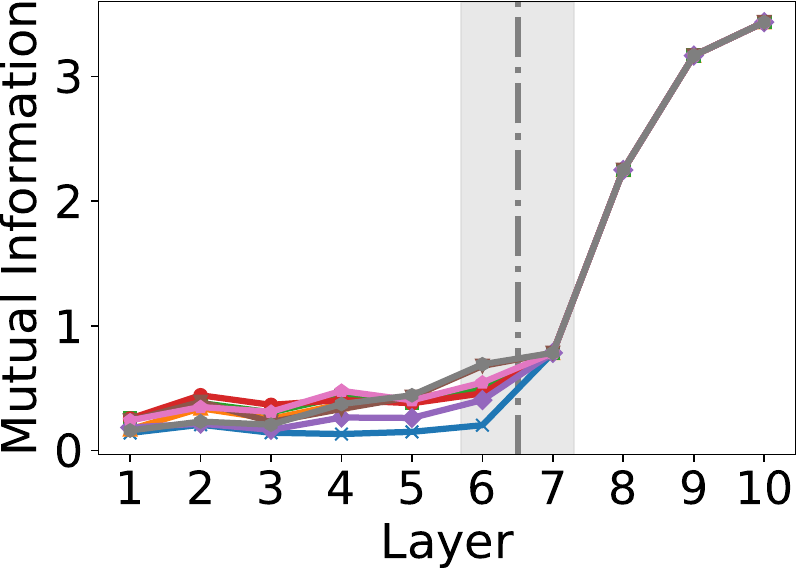}
        \hspace{0.01mm}
        \includegraphics[width=0.19\textwidth]{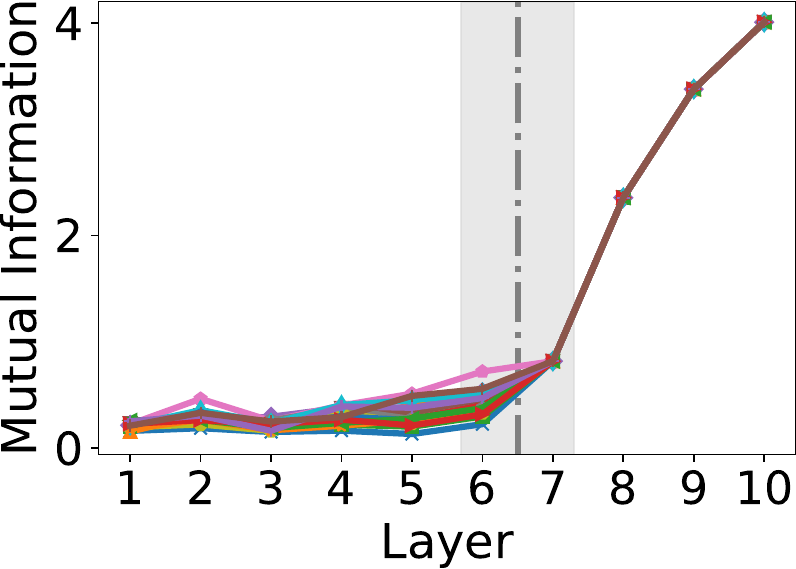}
    }
    \caption{Mutual information between the outputs of each layer and the labels for different datasets and models. In each subfigure, the left side of the gray dotted line corresponds to the bottom model, while the right side corresponds to the top model. Lines of different colors indicate different passive parties. From left to right, the number of passive parties gradually increases.}
    \label{fig:MI}
\end{figure*}

\parahead{Passive parties.} To improve the practical relevance of our study, we extend the experimental setup from traditional two-party scenarios to more realistic multi-party scenarios. The number of passive parties is set to 1, 2, 4, 7, and 14 for the MNIST dataset, and to 1, 2, 4, 8, and 16 for the other datasets. Notably, in two-party scenarios, where there is only one passive party, the overall training process of the VFL model converges to be consistent with that of a traditional, non-distributed or centralized deep neural network. Different passive parties possess distinct features and their own bottom models, which share the same architecture. Each passive party trains its bottom model using its local features and the gradients received from the active party.

\subsection{Observation}

To investigate the respective contributions of the bottom and top models in supervised VFL, we conduct comprehensive experiments across five datasets and five representative model architectures to perform mutual information estimation by~\cite{rossi2006mutual}. The results are presented in Fig.~\ref{fig:MI}. Through these experiments, we obtain the following observations.

\parahead{Observation 1: In VFL scenarios, the mutual information between the outputs of each layer and the labels increases as the layer depth increases.} We observe that for all datasets and model architectures, mutual information between layer outputs and labels exhibits an increasing trend with layer depth. This trend is evident for both bottom models on passive parties and the top model on the active party (special results for MNIST in multi-party scenarios will be discussed in detail in Observation~3). These findings indicate that as the model goes deeper, its ability to capture label-related information is progressively enhanced.

\parahead{Observation 2: The label representation capability of the top model significantly surpasses that of the bottom model, which suggests that the bottom model primarily extracts sample features, while the mapping between features and labels is mainly accomplished by the top model.} We calculate the ratio of the maximum mutual information of the top model to that of the bottom model across the five datasets and five models, finding it ranges from 13.17 to 306.92. This demonstrates the top model's substantially greater capacity for label representation compared to the bottom model. Moreover, we observe that this ratio increases as the number of passive parties grows, indicating that the addition of more passive parties further diminishes the bottom model's ability to represent labels. In addition, the increase in mutual information between the bottom model's layer outputs and the labels with layer depth is much steeper than that of the top model. This further suggests that labels have minimal influence on the bottom model, which primarily learns sample features rather than label information.

\parahead{Observation 3: As the number of passive parties increases, the mutual information between the cut layers of the bottom and top models exhibits a ``transition''.\protect\footnote{We borrow the term ``transition'' from quantum mechanics to describe the abrupt changes in mutual information observed at the cut layers.}} To ensure experimental robustness, we maintain consistent output dimensions for the cut layers of both the bottom and top models throughout our experiments. Across the MNIST, CIFAR-10, and CINIC-10 datasets, we observe that in two-party scenarios~(with only one passive party), the mutual information between the two cut layers remains largely stable. However, as the number of passive parties increases, the mutual information at the cut layers does not increase smoothly. Instead, a distinct and steep increase appears at the top model's cut layer, which we refer to as the ``transition'' phenomenon. We believe the primary reason for this phenomenon is that as the number of passive parties increases, the influence of gradients on the bottom models gradually diminishes. Meanwhile, the distinct features held by each passive party cause the bottom models to learn highly diverse knowledge, resulting in a much weaker association with the labels. To address the insufficient label representation capability of the bottom models, the top model is forced to ``\textbf{compensate}'', resulting in the observed transition in mutual information at the cut layers. Moreover, this compensation may further weaken the label representation capability of the bottom models. For example, on the MNIST dataset, an extreme phenomenon even emerges: as the number of passive parties increases, the mutual information of each layer in bottom models progressively decreases. In addition, we observe that even in the SVHN and GTSRB datasets, where no clear transition occurs, the largest increase in mutual information still appears in the top model. This further highlights the compensatory role of the top model.

\begin{figure}[t!]
    \centering
    \includegraphics[width=0.65\columnwidth]{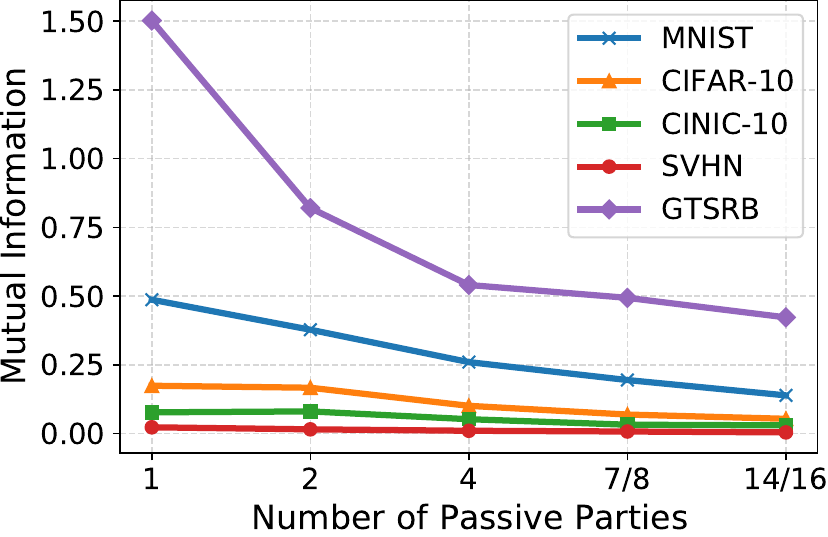}
    \caption{The mutual information between the cut layer of the bottom model and the labels decreases as the number of passive parties increases.}
    \label{fig:MI_PP}
\end{figure}

\parahead{Observation 4: The mutual information at the cut layer of the bottom model exhibits a decreasing trend as the number of passive parties increases.} As shown in Fig.~\ref{fig:MI_PP}, we illustrate the trend of mutual information at the cut layer of the bottom model under varying numbers of passive parties. It is evident that mutual information exhibits a pronounced decline as the number of passive parties increases. This indicates that the bottom model's ability to represent labels progressively diminishes with more passive parties. This finding is consistent with Observation~3, further suggesting that the top model compensates for most of the bottom model's capability to learn labels.

\parahead{} In summary, through extensive experiments across multiple datasets and model architectures, we observe that in VFL scenarios, the bottom models primarily focus on feature extraction, while the top model handles the mapping from features to labels. As the number of passive parties increases, the bottom model becomes less effective at learning label information, prompting the top model to compensate for this deficiency in label representation.

\subsection{Theoretical Analysis}

To support our observations, we provide a theoretical analysis to prove that in VFL scenarios, the mutual information between the outputs of each layer and labels increases with layer depth.

\begin{figure}[t!]
    \centering
    \begin{tikzpicture}
        \node (X1) at(0,0){$X^1$};
        \node (T11) at(1,0){$T_1^1$};
        \node (T1d) at(2,0){$\dots$};
        \node (T1n) at(3,0){$T_n^1$};

        \node (X2) at(0,-0.7){$X^2$};
        \node (T12) at(1,-0.7){$T_1^2$};
        \node (T2d) at(2,-0.7){$\dots$};
        \node (Tn2) at(3,-0.7){$T_n^2$};

        \node (Xd) at(2,-1.27){$\vdots$};

        \node (Xk) at(0,-2.1){$X^k$};
        \node (Tk1) at(1,-2.1){$T_1^k$};
        \node (Tkd) at(2,-2.1){$\dots$};
        \node (Tnk) at(3,-2.1){$T_n^k$};

        \node (S1) at(4,-1.15){$S_1$};
        \node (Sd) at(5,-1.15){$\dots$};
        \node (Sm) at(6,-1.15){$S_m$};
        \node (Y) at(7,-1.15){$Y$};

        \draw[->] (X1) -- (T11);
        \draw[->] (T11) -- (T1d);
        \draw[->] (T1d) -- (T1n);
        \draw[->] (T1n) -- (S1);

        \draw[->] (X2) -- (T12);
        \draw[->] (T12) -- (T2d);
        \draw[->] (T2d) -- (Tn2);
        \draw[->] (Tn2) -- (S1);

        \draw[->] (Xk) -- (Tk1);
        \draw[->] (Tk1) -- (Tkd);
        \draw[->] (Tkd) -- (Tnk);
        \draw[->] (Tnk) -- (S1);

        \draw[->] (S1) -- (Sd);
        \draw[->] (Sd) -- (Sm);
        \draw[->] (Sm) -- (Y);
    \end{tikzpicture}
    \caption{The topology of the lumped Markov chains in VFL.}
    \label{fig:markov}
\end{figure}
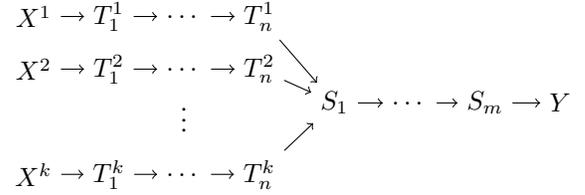

Inspired by works on the information bottleneck~\cite{tishby2000information,tishby2015deep,saxe2019information}, we regard the VFL neural network as the lumped Markov chains~\cite{derisavi2003optimal,geiger2014lumpings}. For each chain $i$, the process can be formalized as $X^i \to T_1^i \to T_2^i \to \dots \to T_n^i \to S_1 \to S_2 \to \dots \to S_m \to Y$, where $S_1$ denotes the lumping point, $X^i$ represents the features held by passive party $i$, ${T_1^i, \dots, T_n^i}$ are the outputs of each layer in the bottom model $f_i(\cdot)$, ${S_1, \dots, S_m}$ are the outputs of each layer in the top model $g(\cdot)$, and $Y$ denotes the labels. Additionally, the bottom model consists of $n$ layers, while the top model consists of $m$ layers. We assume there are $k$ passive parties in total. The topology of the lumped Markov chain is shown in Fig.~\ref{fig:markov}. Our proof consists of three steps.

\parahead{Step \Rmnum{1}: We first prove that for any passive party $i$, the following holds: $I(X^i; Y) \leq I(T_1^i; Y) \leq I(T_2^i; Y) \leq \dots \leq I(T_n^i; Y)$.} Given the two decomposition forms of $I(X^i,T_1^i;Y)$ that
\begin{equation}    
    \begin{aligned}
        I(X^i,T_1^i;Y) &= I(T_1^i; Y) + I(X^i; Y \mid T_1^i)\\
        &= I(X^i; Y) + I(T_1^i; Y \mid X^i),
    \end{aligned}
    \label{eq:step1_1}
\end{equation}
and based on the lumped Markov chains, we know that given $T_1^i$, the conditional independence $Y\indep X^i\mid T_1^i$ is satisfied, i.e., $I(X^i; Y \mid T_1^i) = 0$. Substituting this into the Equation~(\ref{eq:step1_1}) yields
\begin{equation}
    I(T_1^i; Y) = I(X^i; Y) + I(T_1^i; Y \mid X^i).
\end{equation}
Since mutual information is non-negative, $I(T_1^i; Y \mid X^i) \geq 0$, it follows that
\begin{equation}
    I(X^i; Y) \leq I(T_1^i; Y).
\end{equation}
Similarly, for any $j \in \{1, 2, \dots, n-1\}$, we can prove that
\begin{equation}
    I(T_j^i; Y) \leq I(T_{j+1}^i; Y).
\end{equation}
Thus, we conclude that for any passive party $i$, the following holds:
\begin{equation}
    I(X^i; Y) \leq I(T_1^i; Y) \leq I(T_2^i; Y) \leq \dots \leq I(T_n^i; Y).
\end{equation}

\parahead{Step \Rmnum{2}: Next, we prove that for any passive party $i$, it safeties $I(T_n^i;Y) \leq I(S_1;Y)$.} To certify $I(T_n^i;Y) \leq I(S_1;Y)$, it suffices to show that
\begin{equation}
    H(Y) - H(Y \mid T_n^i) \leq H(Y) - H(Y \mid S_1).
\end{equation}
That is equivalent to prove
\begin{equation}
    H(Y \mid T_n^i) \geq H(Y \mid S_1).
\end{equation}
According to the principle that ``conditioning never increases entropy''~\cite{shannon1948mathematical}, it follows that:
\begin{equation}
    \begin{aligned}
        H(Y \mid T_n^i) &\geq H(Y \mid T_n^1,T_n^2,\dots,T_n^k)\\
        &\geq H(Y \mid T_n^1,T_n^2,\dots,T_n^k,S_1).
    \end{aligned}
\end{equation}
Since $Y\indep T_n^1,T_n^2,\dots,T_n^k\mid S_1$, we have
\begin{equation}
    H(Y \mid T_n^1,T_n^2,\dots,T_n^k,S_1) = H(Y \mid S_1).
\end{equation}
Combining the above equations, $H(Y \mid T_n^i) \geq H(Y \mid S_1)$ holds. Therefore, for any passive party $i$, it safeties $I(T_n^i;Y) \leq I(S_1;Y)$.

\parahead{Step \Rmnum{3}: Finally, we prove that $I(S_1; Y) \leq I(S_2; Y) \leq \dots \leq I(S_m; Y)$ holds.} Since $S_1 \to S_2 \to \dots \to S_m$ forms a Markov chain, following the data processing inequality~(DPI) theory~\cite{beaudry2011intuitive}, we obtain
\begin{equation}
    I(S_1; Y) \leq I(S_2; Y) \leq \dots \leq I(S_m; Y).
\end{equation}

In summary, by combining the results from the three steps, we prove that for any passive party $i$,
\begin{equation}
    \begin{aligned}
    I(X^i; Y) &\leq I(T_1^i; Y) \leq I(T_2^i; Y) \leq \dots \leq I(T_n^i; Y)\\ &\leq I(S_1; Y) \leq I(S_2; Y) \leq \dots \leq I(S_m; Y).        
    \end{aligned}
\end{equation}
In other words, the mutual information between the outputs of each layer and labels increases as the layer depth increases. This aligns with our experimental observations in Fig.~\ref{fig:MI} and provides theoretical support for them.


\section{Attackers Can Be Victims Too}
\label{sec:attackers_victims}

Many existing LIA methods are built on an intuitive assumption in VFL: embeddings generated by a well-trained bottom model exhibit a strong mapping to the corresponding labels. Thus, LIA can be achieved either through clustering properties of these embeddings or by fine-tuning the bottom model using a small amount of labeled auxiliary data. However, based on our findings in the previous section regarding the distinct functional roles of bottom and top models, we believe that this intuitive assumption may have misled existing LIA research. In fact, the apparent success of existing LIA methods is likely due to a coincidental alignment between the distributions of sample features and labels, which inadvertently strengthens the mapping between embeddings and labels. Therefore, in this section, we introduce task reassignment to reveal the vulnerability of existing LIAs based on this intuitive embedding-label assumption.

\begin{table*}[t!]
    \centering
    \small
    \setlength{\tabcolsep}{4.5pt}
    \caption{Performance of LIAs under newly assigned tasks}
    \label{tab:reassignment}
    \begin{threeparttable}
        \begin{tabular}{cccccccccc}
            \toprule
            \multirow{2}{*}[-0.3em]{LIA} & \multirow{2}{*}[-0.3em]{Dataset} & \multicolumn{2}{c}{Original Task} & \multicolumn{2}{c}{New Task 1} & \multicolumn{2}{c}{New Task 2} & \multicolumn{2}{c}{New Task 3} \\
            \cmidrule{3-10}
            & & Attack Acc. & MTA & Attack Acc. & $\Delta$MTA & Attack Acc. & $\Delta$MTA & Attack Acc. & $\Delta$MTA \\
            \midrule
            \multirow{5}{*}{Cluster} & MNIST & 85.54$\pm$6.54\% & 97.10$\pm$0.06\% & 37.48$\pm$3.50\% & -0.002 & 25.56$\pm$4.02\% & -0.000 & 17.00$\pm$4.27\% & 0.003 \\
            & CIFAR-10 & 18.66$\pm$3.92\% & 60.44$\pm$0.95\% & 17.29$\pm$1.37\% & -0.004 & 11.62$\pm$3.67\% & 0.004 & 10.89$\pm$0.85\% & -0.006 \\
            & CINIC-10 & 13.78$\pm$0.68\% & 74.82$\pm$1.34\% & 12.32$\pm$1.28\% & -0.012 & 11.79$\pm$1.95\% & 0.004 & 10.79$\pm$0.65\% & -0.005 \\
            & SVHN & 20.50$\pm$6.21\% & 90.11$\pm$0.90\% & 11.46$\pm$1.57\% & 0.001 & 9.70$\pm$0.31\% & -0.001 & 9.56$\pm$0.62\% & -0.004 \\
            & GTSRB & 36.33$\pm$3.21\% & 73.71$\pm$3.87\% & 12.69$\pm$1.65\% & -0.008 & 11.47$\pm$2.14\% & -0.001 & 4.42$\pm$2.06\% & -0.012 \\
            \midrule
            \multirow{5}{*}{Completion} & MNIST & 93.90$\pm$1.55\% & 97.08$\pm$0.08\% & 47.07$\pm$0.30\% & -0.001 & 27.26$\pm$1.39\% & 0.000 & 18.66$\pm$0.64\% & 0.004 \\
            & CIFAR-10 & 29.62$\pm$1.98\% & 60.48$\pm$2.37\% & 23.52$\pm$1.55\% & 0.006 & 17.34$\pm$0.41\% & 0.009 & 15.25$\pm$0.72\% & 0.003 \\
            & CINIC-10 & 26.83$\pm$1.04\% & 72.47$\pm$1.93\% & 21.54$\pm$0.42\% & -0.014 & 16.55$\pm$1.45\% & -0.015 & 16.28$\pm$0.60\% & -0.003 \\
            & SVHN & 64.53$\pm$2.60\% & 88.93$\pm$2.84\% & 31.76$\pm$1.52\% & 0.015 & 19.86$\pm$1.15\% & 0.019 & 14.53$\pm$1.36\% & -0.018 \\
            & GTSRB & 64.65$\pm$3.61\% & 72.57$\pm$1.93\% & 28.55$\pm$2.80\% & 0.005 & 20.98$\pm$1.84\% & 0.009 & 7.29$\pm$0.81\% & 0.012 \\
            \bottomrule
        \end{tabular}
        \begin{tablenotes}
            \footnotesize
            \item[*] ``Acc.'' denotes accuracy. The $\Delta$MTA values are calculated relative to the original task and are kept minimal to ensure a fair evaluation of attack accuracy.
        \end{tablenotes}
    \end{threeparttable}
\end{table*}

\subsection{Threat Model}

The threat model describes the security assumptions of LIAs, which is consistent with those in previous works~\cite{fu2022label,kariyappa2023exploit,liu2024similarity,zou2024defending}.

\parahead{Attackers' Objective.} As an attacker, the passive party aims to infer the private labels of the active party. This attack can occur in both the training phase and the inference phase, but it is typically not carried out in the inference phase alone.

\parahead{Attackers' Capability.} We assume that any passive party can launch the attack independently, with no collusion between attackers. Additionally, we consider the attackers to be \textit{honest-but-curious}. This means they will follow the VFL training protocol while attempting to exploit any available information for their attack. Specifically, the attacker can collect embeddings generated by the bottom model during both the training and inference phases, as well as gradients backpropagated by the active party at any stage of training. In the inference phase, the attacker can leverage all available prior knowledge to infer labels. Notably, the attacker cannot obtain any information about the top model, including its structure, size, and parameters.

\parahead{Attackers' Knowledge.} The attacker possesses private features and a bottom model, along with knowledge of the training objectives (e.g., label categories and number of label classes). Before training begins, all participants complete sample alignment and obtain a global index for each training sample. During the training phase, the knowledge obtained includes models at different iterations and their corresponding embeddings and gradients. In the inference phase, it includes the fixed bottom model and its generated embeddings. Typically, the attacker can also acquire a small amount of auxiliary data through public purchase or negotiation, such as labeled data or label distributions.

\subsection{Settings}

\paraheadtop{Datasets and Models.} We utilize the same five datasets and model architectures as described in Section~\ref{sec:model_comp_setup}.

\parahead{Attacks.} We conduct experiments using two representative types of LIAs described in Section~\ref{sec:lia}: LIA with cluster~\cite{liu2024similarity} and LIA with model completion~\cite{fu2022label}. Both approaches are based on the intuitive embedding-label assumption.

\parahead{Task Reassignment.} To challenge the foundational assumption of existing LIA methods, we introduce the task reassignment. By assigning new training objectives to the original prediction tasks, this approach effectively eliminates the natural correspondence between features and original labels within the dataset. It is important to note that this natural alignment typically stems from the quantitative correspondence between feature categories and label categories, rather than from a correspondence in their names or content. For example, in the MNIST dataset, the features correspond to 10 distinct handwritten digits, naturally forming a 10-class distribution. If we only change the label content without altering the number of label categories, the label distribution remains 10-class, and the alignment between features and labels in terms of distribution cannot be disrupted. Therefore, to effectively break this alignment, we introduce three distinct new classification tasks, each with progressively fewer label categories compared to the original task. This implies an increasing disparity between the sample feature distribution and the label distribution. By doing so, we attempt to reveal the vulnerability of existing embedding-label-assumption-based LIAs by comparing their performance on the original tasks and these new tasks. When designing these three new tasks, we carefully consider their reasonableness and practical relevance to ensure clear real-world significance. For example, the objective of New Task 3 on the MNIST dataset is to classify handwritten digits by parity. In New Task 1, the original CIFAR-10 labels are aggregated as follows: ``airplane'' and ``ship'' to ``transport''; ``automobile'' and ``truck'' to ``vehicle''; ``bird'' and ``frog'' to ``small animal''; ``cat'' and ``dog'' to ``pet''; and ``deer'' and ``horse'' to ``ungulate''. Moreover, we also strive to balance the amount of samples across different new labels. The specific definitions of these three new tasks for each dataset are detailed in Appendix~A.

\parahead{Evaluation Metrics.} We evaluate the performance of LIAs using \textit{attack accuracy}, consistent with previous works~\cite{fu2022label,kariyappa2023exploit,liu2024similarity,zou2024defending}. Attack accuracy is defined as the ratio of correctly inferred labels to the total number of samples, with higher values indicating better attack performance. It is worth noting that our task reassignment reduces the number of label categories in the new tasks compared to the original task. As a result, the random guessing accuracy for label inference increases, effectively raising the baseline for LIA performance. To ensure a fair comparison of LIA performance across different tasks, we calculate the \textit{lift}~\cite{brin1997dynamic,shani2011evaluating} in attack accuracy relative to random guessing and map this relative performance back to the original task's metric space~\cite{cohen1960coefficient}. For example, in MNIST, the random guessing accuracy for label inference on the original 10-class classification task is 10\%, while on the parity classification task it is 50\%. Suppose an LIA achieves 25\% attack accuracy on the original task and 40\% on the parity task. It is evident that although the attack accuracy appears to improve from 25\% to 40\%, the 40\% is actually below the 50\% random guessing baseline for the parity task, indicating a substantial decline in real label inference capability. Therefore, when comparing performance to the original task~(using its random guessing accuracy as the baseline), this attack accuracy should be mapped as $40\%\div 50\%\times 10\%=8\%$ to reflect the fact that its performance is actually worse than random guessing. In addition, we also use the main task accuracy~(MTA) to assess the predictive performance of the VFL model.

\subsection{Results and Analysis}
\label{sec:task_reassignment}

We first evaluate the attack performance of LIAs based on the embedding-label assumption across different classification tasks, demonstrating that their high attack accuracy is essentially an illusion.

To control for variables and ensure that the overall VFL model exhibits comparable label prediction capabilities across different tasks, we evaluate the performance of LIAs in each task while maintaining essentially consistent MTA levels. The results for two-party VFL are presented in Table~\ref{tab:reassignment}, while the results for multi-party VFL are provided in Table~\Rmnum{2} of Appendix~B. It can be observed that all training tasks achieve similar MTA values, indicating that the overall VFL model can still effectively learn label information on the new tasks. However, LIAs implemented using these well-trained bottom models and their generated embeddings exhibit significantly degraded attack performance on the new tasks. This suggests that the bottom models do not genuinely learn label information during training, but instead primarily capture the structural features of the data. This finding corroborates our conclusion in Section~\ref{sec:model_compensation}: in VFL, labels are primarily learned by the top model, while the bottom models focus more on feature extraction.

In addition, as shown in the table, across all five datasets, the attack accuracy of two types of LIAs gradually declines as the discrepancy between the new tasks and the original task increases. This phenomenon primarily occurs because reducing the number of label categories in the new tasks disrupts the original alignment between the distributions of features and labels. In the original task, there is a relatively consistent correspondence between sample features and labels. Thus, even though the bottom model in VFL mainly learns sample features rather than labels, the embeddings generated from these features typically align with the labels, enabling strong attack performance for LIAs. However, as the task objective shifts from intuitive classification to more abstract classification, the feature structure learned by the VFL bottom model struggles to establish an effective mapping with the new labels, resulting in a significant decline in LIA performance.

Moreover, we observe that as the discrepancy between the label distribution of new tasks and the feature distribution of samples increases, the performance of the two LIAs even falls below the level of random guessing on certain datasets, such as SVHN and GTSRB. This indicates a complete failure of their label inference capabilities. Since this phenomenon is especially pronounced on real-world datasets, we infer that LIAs based on the embedding-label assumption may experience even greater declines in attack performance when faced with more complex and abstract tasks in practical applications.

\begin{figure*}[t!]
    \centering
    \subfloat[MNIST]{\includegraphics[width=\textwidth]{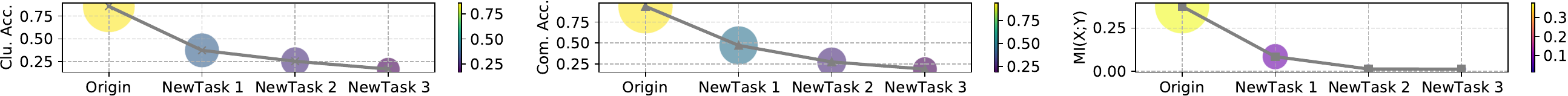}} \\
    \subfloat[CIFAR-10]{\includegraphics[width=\textwidth]{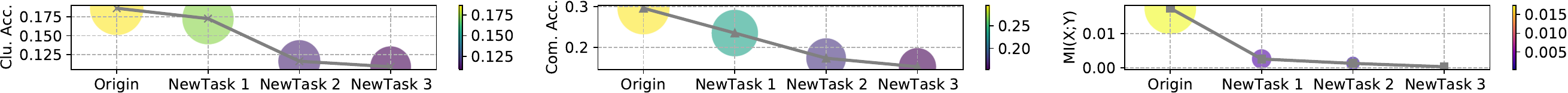}} \\
    \subfloat[CINIC-10]{\includegraphics[width=\textwidth]{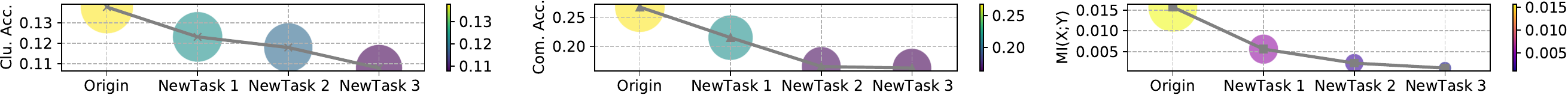}} \\    
    \subfloat[SVHN]{\includegraphics[width=\textwidth]{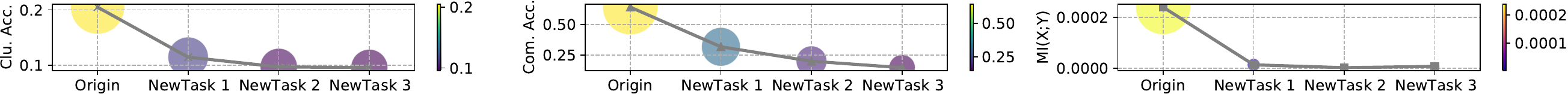}} \\
    \subfloat[GTSRB]{\includegraphics[width=\textwidth]{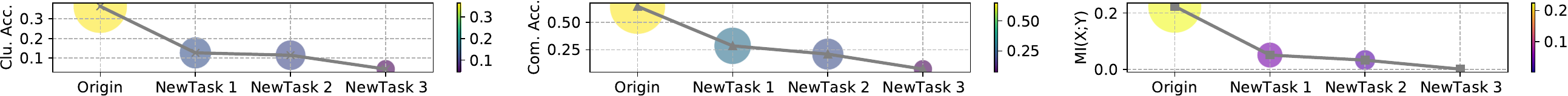}} \\
    \caption{Trends of attack accuracy and mutual information under different tasks. The first column presents the attack accuracy of LIA with cluster method. The second column presents the attack accuracy of LIA with model completion method. The third column presents the mutual information between the dataset features and the labels.}
    \label{fig:reassignment_trend}
\end{figure*}

To further investigate the impact of the correlation between sample features and labels on LIA performance, we first calculate the mutual information between features and labels across different tasks to quantify the variation in the alignment between label and sample feature distributions. We then compare the trends in mutual information across tasks with the trends in attack accuracy for the two LIAs. The results are presented in Fig.~\ref{fig:reassignment_trend}. As shown in the figure, for all five datasets, as the task changes from the original to the new tasks, the mutual information between sample features and labels consistently decreases. This indicates a weakening alignment between the label distribution and the sample feature distribution. Notably, we observe that the trends in mutual information closely mirror the trends in attack accuracy for both LIAs. This suggests that the performance of these LIAs is highly dependent on the degree of alignment between sample features and labels. When this alignment is strong, as in the original task, these LIAs can achieve high attack accuracy. However, as the alignment weakens in the new tasks, their attack performance deteriorates significantly.

In summary, through task reassignment experiments, we demonstrate that existing LIAs based on the embedding-label assumption are vulnerable. Their high attack accuracy is largely an illusion, contingent on a strong natural alignment between sample features and labels. When this alignment is disrupted, the model compensation of VFL further reinforces the bottom model's tendency to focus on learning feature extraction rather than label representations. As a result, LIA performance drops sharply, potentially even falling below the random guessing level. Based on these findings, we caution against overreliance on intuitive assumptions that may not hold in practice, urge careful evaluation of the effectiveness of existing LIA methods, and encourage future research to develop more robust LIA approaches.

\subsection{Broader Impacts}

Given that LIA is a unique attack specific to the VFL scenario, our exposure of the vulnerability in existing LIAs may have broader impacts for the entire VFL security community. For instance, backdoor attacks~\cite{bai2023villain,shen2025label,chen2025backdoor,liu2026trigger}, which are prevalent threats in distributed machine learning systems, can also introduce considerable risks in VFL scenarios. These attacks allow passive or active attackers to implant backdoors into models during training, causing label misclassification during inference. Backdoor attacks are typically categorized as either untargeted, where the model outputs any incorrect label for trigger-containing samples, or targeted, where the incorrect label must match a specific target category. In VFL, when the passive party acts as the attacker, since it possesses only sample features without corresponding labels, executing a targeted backdoor attack requires prior label inference. As a result, label inference attacks often serve as a prerequisite for backdoor attacks in VFL. Therefore, our findings regarding the vulnerabilities of existing LIAs could also have significant impacts for the feasibility and effectiveness of backdoor attacks in VFL. If existing LIAs are less effective than previously believed, this could limit the ability of attackers to successfully implement targeted backdoor attacks, thereby enhancing the overall security of VFL systems. Consequently, our work not only challenges current assumptions about LIA effectiveness but also contributes to a broader understanding of security vulnerabilities in VFL.

In addition, in our practical experiments, we also investigate the impact of the actual roles of bottom and top models on gradient-based LIAs. Existing gradient-based LIAs are typically categorized into two types: (1) attackers who can obtain gradients associated with the top model's logits and infer the label category based on the gradient signs~\cite{wainakh2021label,wainakh2021user}; and (2) attackers who exploit the fact that data from different label categories generally cause the model to optimize in different directions, enabling label inference by classifying the gradients~\cite{liu2024similarity}. We conduct experiments on both types of gradient-based attacks. The results indicate that the first attack is virtually unaffected, as it can directly access gradients related to label predictions. However, this attack requires stricter security assumptions and is generally restricted to binary classification rather than more realistic multi-class tasks, which limits its practicality in real-world scenarios. The second attack typically occurs during the very early stages of model training, when gradient changes are most pronounced, allowing attackers to identify classification features from these gradients and perform LIAs. As training progresses and stabilizes, the differences between gradients of different labels gradually diminish, making LIAs significantly more difficult to execute. Although the implementation of this attack is largely independent of the embedding-label assumption discussed in the paper, we include experiments on this method for completeness. We find that the attack accuracy of this gradient-based LIA does not vary significantly whether the attack is performed during the early training phase or after the model is trained. Furthermore, even in relatively simple two-party scenarios and on the MNIST dataset, the accuracy of such attacks remains very low and sometimes even below random guessing. These experimental results are consistent with those reported by~\cite{liu2025attackers}. In summary, our findings suggest that gradient-based LIAs may also be less effective than previously believed, particularly in more complex and realistic VFL scenarios. This further underscores the importance of re-evaluating the assumptions underlying LIA methods and encourages the development of more robust attack strategies that can withstand the unique challenges posed by VFL.

\begin{table*}[t!]
    \centering
    \small
    \setlength{\tabcolsep}{4.4pt}
    \caption{The impact of cut layer position on MTA and attack accuracy}
    \label{tab:defense_move}
    \begin{threeparttable}
        \begin{tabular}{cccccccccc}
            \toprule
            \multirow{2}{*}[-0.3em]{LIA} & \multirow{2}{*}[-0.3em]{Dataset} & \multicolumn{2}{c}{Cut Layer Position: -2} & \multicolumn{2}{c}{Cut Layer Position: -3} & \multicolumn{2}{c}{Cut Layer Position: -4} & \multicolumn{2}{c}{Cut Layer Position: -5} \\
            \cmidrule{3-10}
            & & Attack Acc. & MTA & Attack Acc. & $\Delta$MTA & Attack Acc. & $\Delta$MTA & Attack Acc. & $\Delta$MTA \\
            \midrule
            \multirow{5}{*}{Cluster} & MNIST & 94.72$\pm$4.70\% & 97.04$\pm$0.09\% & 88.25$\pm$6.68\% & \textcolor{mygreen}{0.002} & 86.83$\pm$6.22\% & \textcolor{mygreen}{0.001} & 85.54$\pm$6.54\% & \textcolor{mygreen}{0.001} \\
            & CIFAR-10 & 37.93$\pm$7.35\% & 60.09$\pm$1.79\% & 28.56$\pm$5.90\% & \textcolor{mygreen}{0.010} & 19.44$\pm$3.36\% & \textcolor{mygreen}{0.009} & 18.66$\pm$3.92\% & \textcolor{mygreen}{0.004} \\
            & CINIC-10 & 44.31$\pm$10.81\% & 73.35$\pm$2.78\% & 24.35$\pm$7.47\% & \textcolor{mygreen}{0.007} & 16.96$\pm$4.41\% & \textcolor{myred}{-0.003} & 13.78$\pm$0.68\% & \textcolor{mygreen}{0.015} \\
            & SVHN & 72.96$\pm$8.20\% & 89.80$\pm$0.61\% & 40.61$\pm$10.24\% & \textcolor{mygreen}{0.002} & 30.10$\pm$4.20\% & \textcolor{mygreen}{0.001} & 20.50$\pm$6.21\% & \textcolor{mygreen}{0.003} \\
            & GTSRB & 55.50$\pm$4.67\% & 72.52$\pm$2.48\% & 54.58$\pm$5.22\% & \textcolor{mygreen}{0.000} & 49.59$\pm$5.02\% & \textcolor{mygreen}{0.003} & 36.33$\pm$3.21\% & \textcolor{mygreen}{0.012} \\
            \midrule
            \multirow{5}{*}{Completion} & MNIST & 96.46$\pm$0.37\% & 97.05$\pm$0.10\% & 96.19$\pm$0.32\% & \textcolor{mygreen}{0.001} & 95.18$\pm$0.85\% & \textcolor{mygreen}{0.000} & 93.90$\pm$1.55\% & \textcolor{mygreen}{0.000} \\
            & CIFAR-10 & 43.22$\pm$1.92\% & 60.47$\pm$0.99\% & 40.64$\pm$4.17\% & \textcolor{mygreen}{0.005} & 33.64$\pm$2.49\% & \textcolor{myred}{-0.001} & 29.20$\pm$4.49\% & \textcolor{mygreen}{0.003} \\
            & CINIC-10 & 52.75$\pm$2.78\% & 75.39$\pm$1.69\% & 41.23$\pm$1.48\% & \textcolor{mygreen}{0.006} & 30.76$\pm$1.78\% & \textcolor{mygreen}{0.000} & 26.48$\pm$1.54\% & \textcolor{myred}{-0.001} \\
            & SVHN & 83.94$\pm$2.98\% & 88.84$\pm$2.02\% & 81.04$\pm$3.90\% & \textcolor{mygreen}{0.004} & 73.53$\pm$2.83\% & \textcolor{mygreen}{0.006} & 64.53$\pm$2.60\% & \textcolor{mygreen}{0.001} \\
            & GTSRB & 79.82$\pm$3.26\% & 72.25$\pm$3.08\% & 78.49$\pm$3.31\% & \textcolor{mygreen}{0.004} & 75.39$\pm$3.84\% & \textcolor{mygreen}{0.027} & 64.65$\pm$3.61\% & \textcolor{mygreen}{0.003} \\
            \bottomrule
        \end{tabular}
        \begin{tablenotes}
            \footnotesize
            \item[*] The cut layer position is counted from the last layer of the entire model. For example, ``-2'' indicates that the cut layer is the second-to-last layer of the entire model. ``Acc.'' denotes the accuracy. The $\Delta$MTA values are calculated relative to the values at ``Cut Layer Position: -2''. Green indicates an increase, while red indicates a decrease. A higher MTA value indicates better predictive capability of the model.
        \end{tablenotes}
    \end{threeparttable}
\end{table*}


\section{Deepen Your Network}
\label{sec:deepen_network}

We highlight in Section~\ref{sec:attackers_victims} that existing LIAs based on the intuitive embedding-label assumption have inherent limitations. However, the natural mapping between features and labels for training is hard to avoid. Prior research~\cite{shwartzziv2017opening} also shows that, under supervised learning, the label representation capability of each layer generally increases with more training epochs. Therefore, the bottom model may still capture some label information. These factors indicate that such LIAs remain a potential threat in VFL. Our observations in Section~\ref{sec:model_compensation} demonstrate that layers closer to the top model exhibit stronger label representation in VFL. Motivated by this insight, we believe an intuitive approach to mitigating such LIAs is to deepen the top model by increasing the proportion of top layers in the entire model. This shifts the bottom layers earlier in the architecture, reducing their label representation capability and thereby providing effective defense. In this section, we validate this idea and explore its potential implications.

\subsection{Providing Defensive Effects}

\begin{figure}[t!]
    \centering
    \includegraphics[width=\columnwidth]{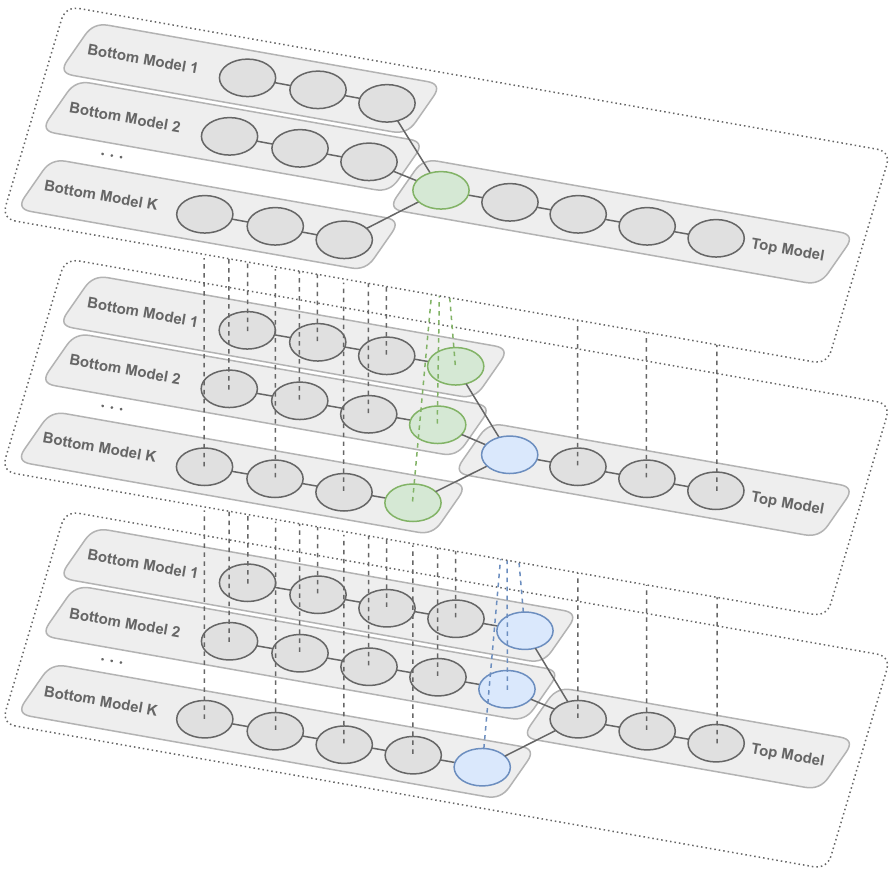}
    \caption{An illustration of moving the cut layer position. Each layer of the model is depicted as an ellipse. We keep the total number of layers in the VFL model unchanged and only adjust the position of the cut layer.}
    \label{fig:move_cut_layer}
\end{figure}

To assess whether moving the cut layer forward enhances defense, we conduct experiments by adjusting the cut layer position closer to the bottom model while keeping the total number of layers unchanged, as shown in Fig.~\ref{fig:move_cut_layer}. This effectively increases the number of top layers and reduces the number of bottom layers, thereby weakening the bottom model's label representation capability. The results for two-party VFL are presented in Table~\ref{tab:defense_move}, while the results for multi-party VFL are detailed in Table~\Rmnum{3} of Appendix~B.

From the table, we observe that, under identical hyperparameters and training epochs, both LIAs show a consistent decline in attack accuracy across all five datasets as the cut layer is moved forward. For most datasets, advancing the cut layer by three layers reduces attack accuracy by more than 50\%, with some datasets experiencing drops of around 70\%---nearly reaching the level of random guessing. Notably, for the complex dataset CINIC-10 and real-world dataset SVHN, moving the cut layer forward by just one layer results in nearly a 50\% reduction in attack accuracy. This suggests that applying this technique in real-world scenarios may lead to even stronger defensive outcomes. Above results demonstrate that reducing the proportion of bottom model layers by advancing the cut layer position can effectively defend against LIAs based on the embedding-label assumption.

In addition to the substantial decline in attack accuracy, we also observe a slight improvement in the MTA metric across all five datasets as the cut layer is moved forward. This indicates that increasing the proportion of top layers not only mitigates LIA threats but also enhances the overall predictive performance of the VFL model. We believe this improvement may be attributed to the model compensation phenomenon of the top model in VFL, which endows it with stronger capacity for learning label representations, thereby better complementing the feature extraction role of the bottom model. Increasing the number of layers in the top model enhances its capacity to represent labels and improves the overall predictive performance of the model, thereby increasing the MTA. Furthermore, moving the cut layer forward enables earlier aggregation of the distinct features held by passive parties, allowing the top model to learn global feature information sooner and establish a more comprehensive mapping to the labels. This also contributes to the improvement in MTA.

Overall, moving the cut layer position forward to increase the proportion of top layers not only effectively defends against LIAs but also enhances the predictive performance of the VFL model.

\begin{figure}[t!]
    \centering
    \includegraphics[width=\columnwidth]{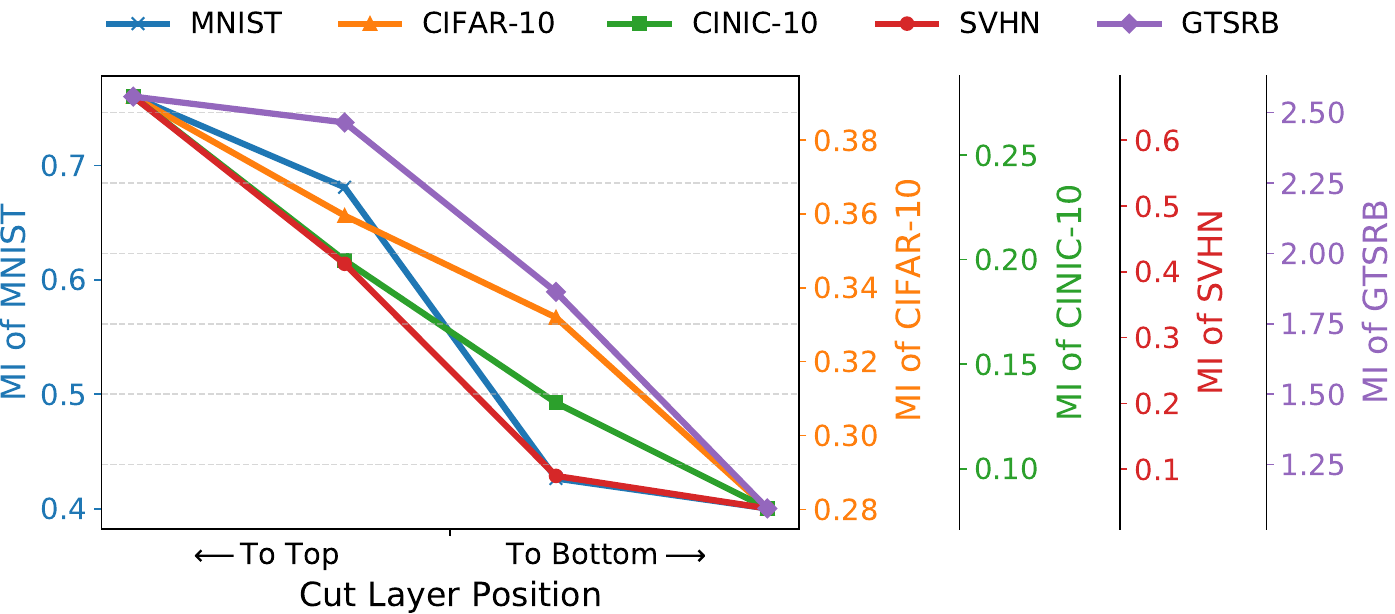}
    \caption{Mutual information decreases when the cut layer is moved closer to the bottom of the model.}
    \label{fig:move_cut_layer_MI}
\end{figure}

To further investigate the reasons behind the effectiveness of this defense technique, we calculate the mutual information between the bottom model's output embeddings and the labels as the cut layer is moved forward. The results are presented in Fig.~\ref{fig:move_cut_layer_MI}. We observe that, across all five datasets, mutual information values consistently decrease as the cut layer advances, aligning with the decline in attack accuracy. This indicates that the reduction in mutual information between the bottom model's outputs and the labels caused by advancing the cut layer is the primary reason for the defense's effectiveness. In other words, moving the cut layer forward effectively weakens the label representation capability of the bottom model, thereby undermining the foundation of LIAs based on the embedding-label assumption. This finding further validates the effectiveness of our proposed defense technique.

\subsection{Enhancing Other Defenses}

Beyond defending against LIAs, moving the cut layer forward can also enhance the effectiveness of other existing defense strategies. To demonstrate this, we evaluate five representative defense strategies, including three general defenses~\cite{abadi2016deep,fang2023improved,fu2022label} applicable to a broad range of VFL scenarios and two specific defenses~\cite{zou2024defending,qiu2023defending} designed specifically for LIAs. These defense strategies are detailed below.

\parahead{Differential Privacy.} Differential privacy~(DP)~\cite{dwork2006differential} is a widely used defense mechanism in VFL because it provides privacy guarantees and quantifies privacy simultaneously. It works by adding carefully calibrated noise to embeddings or gradients exchanged during training to obscure sensitive information. The DP-SGD algorithm~\cite{abadi2016deep} is the most representative implementation of DP in federated learning and has been further extended in recent works~\cite{bai2023villain,liu2024similarity}.

\parahead{Gradient Clipping.} Gradient clipping~\cite{fang2023improved,mao2023differential} is another general defense strategy that limits the magnitude of gradients exchanged during training by clipping those that are excessively large or small. By constraining gradients in this way, it reduces the risk of information leakage through gradient updates. In our implementation, we use norm-based gradient clipping for this defense.

\parahead{Gradient Compression.} Gradient compression~\cite{fu2022label,arazzi2023blindsage,zou2024defending} reduces the amount of information transmitted during training by compressing gradients, thereby limiting the information accessible to potential attackers. For LIAs, gradient compression can decrease the label information learned by bottom models to achieve the defense objective.

\begin{figure*}[t!]
    \centering
    \includegraphics[width=0.6\textwidth]{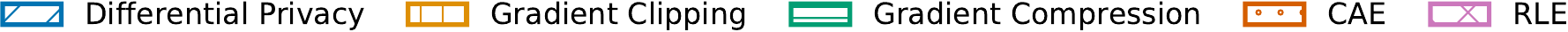}\\
    \subfloat[Cluster, CIFAR-10]{\includegraphics[width=0.24\textwidth]{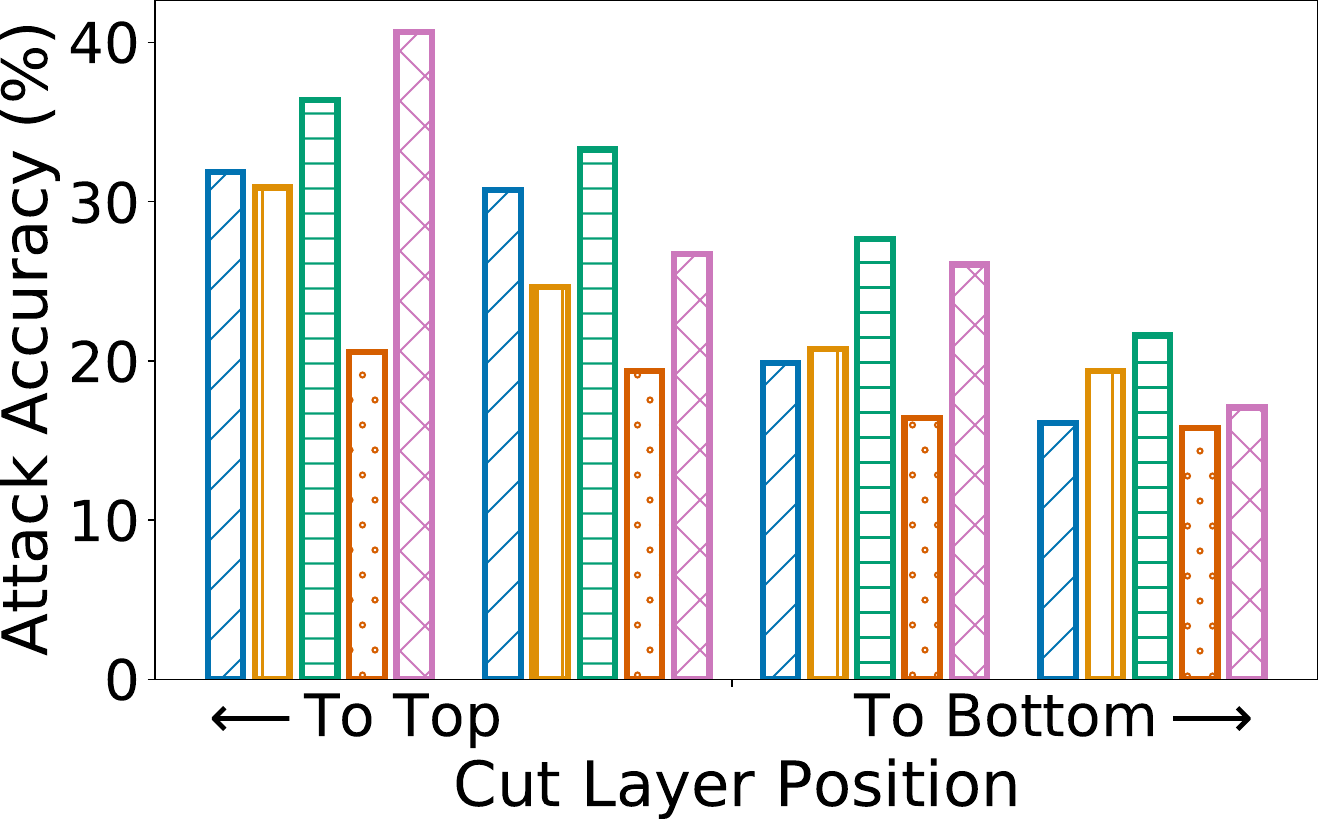}}
    \hspace{1mm}
    \subfloat[Cluster, CINIC-10]{\includegraphics[width=0.24\textwidth]{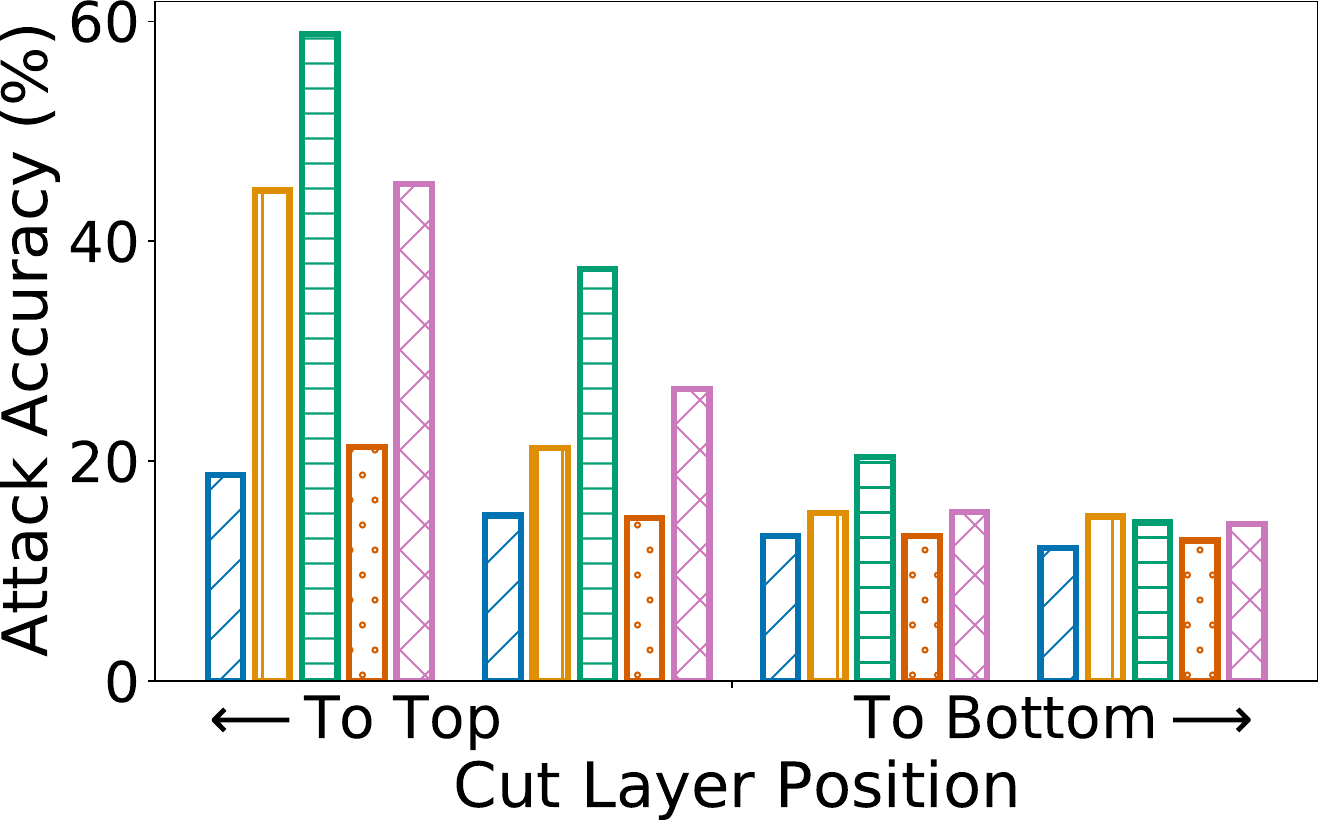}}
    \hspace{1mm}
    \subfloat[Cluster, SVHN]{\includegraphics[width=0.24\textwidth]{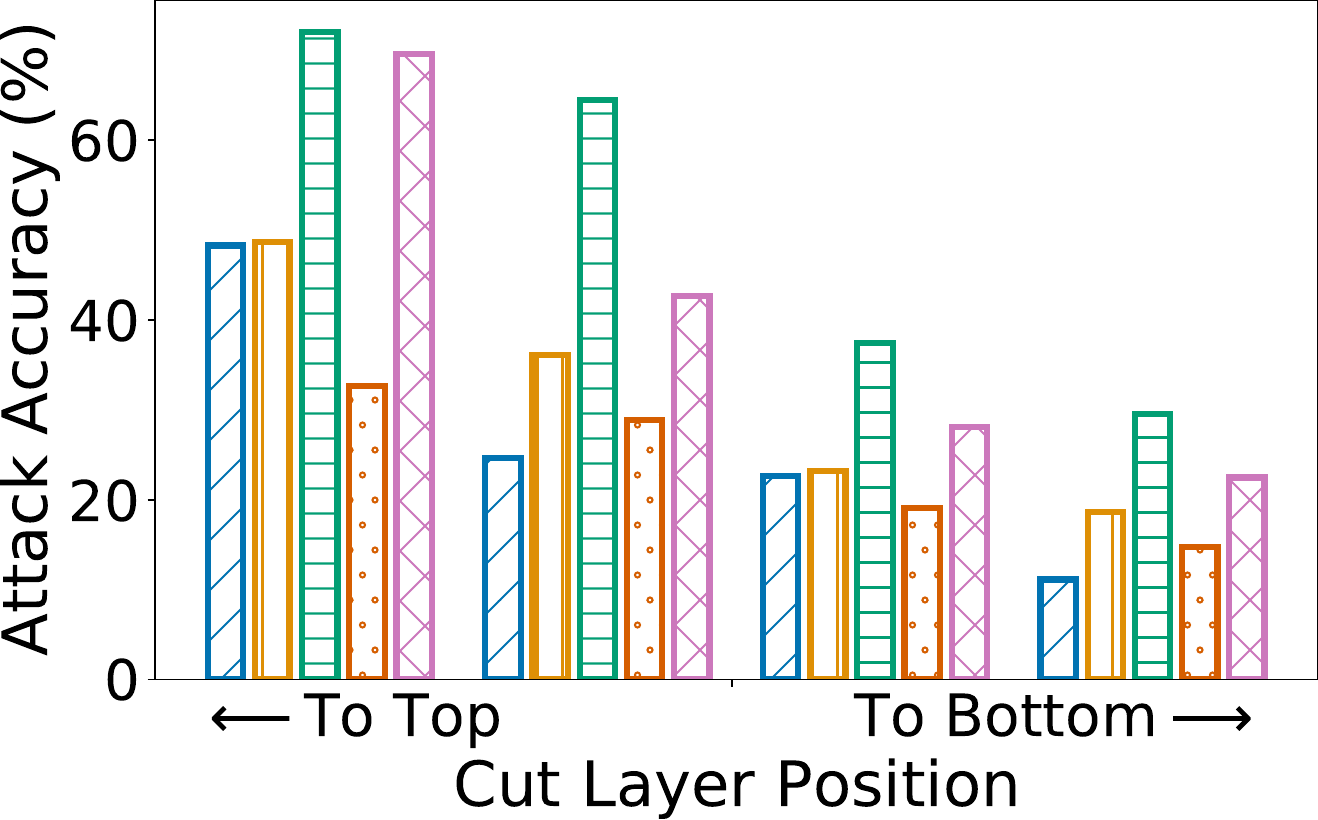}}
    \hspace{1mm}
    \subfloat[Cluster, GTSRB]{\includegraphics[width=0.24\textwidth]{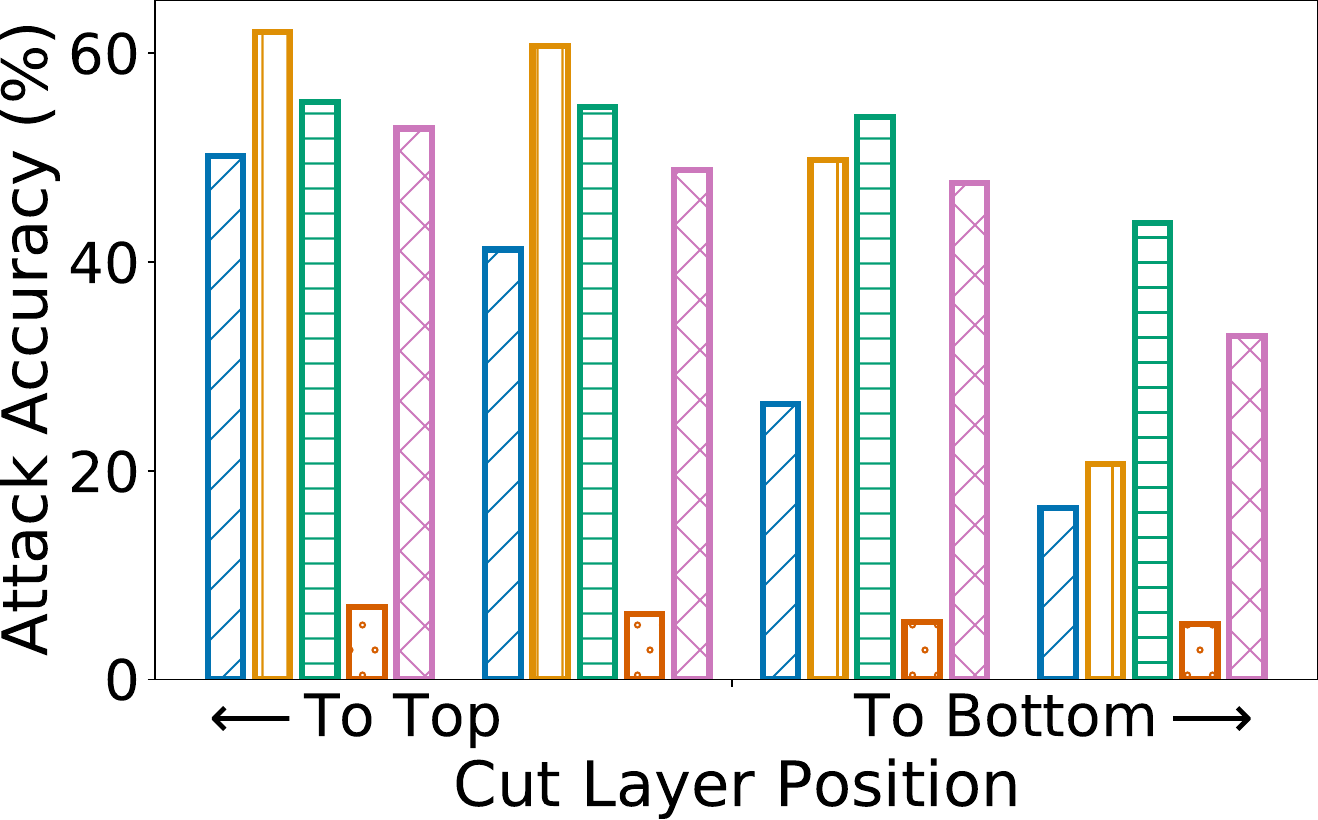}}\\
    \subfloat[Completion, CIFAR-10]{\includegraphics[width=0.24\textwidth]{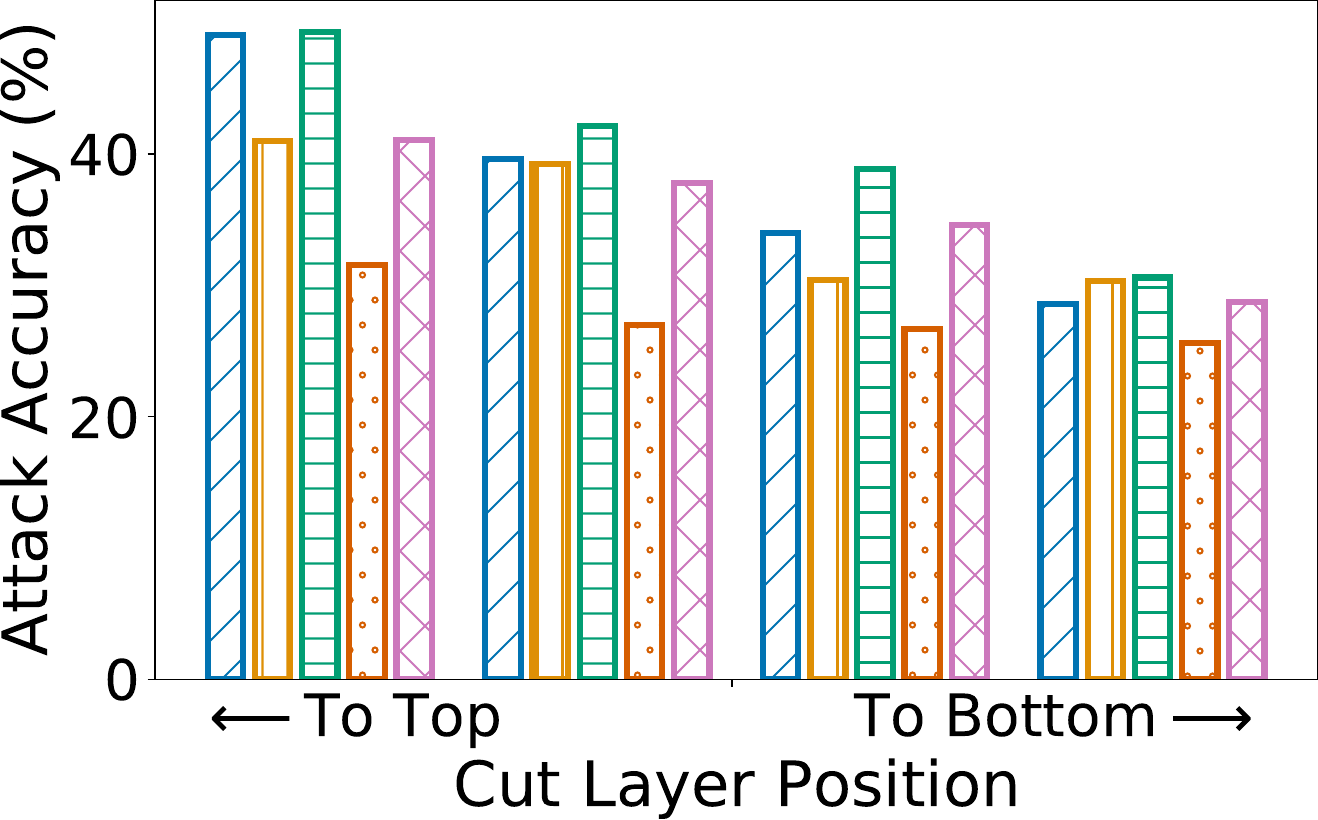}}
    \hspace{1mm}
    \subfloat[Completion, CINIC-10]{\includegraphics[width=0.24\textwidth]{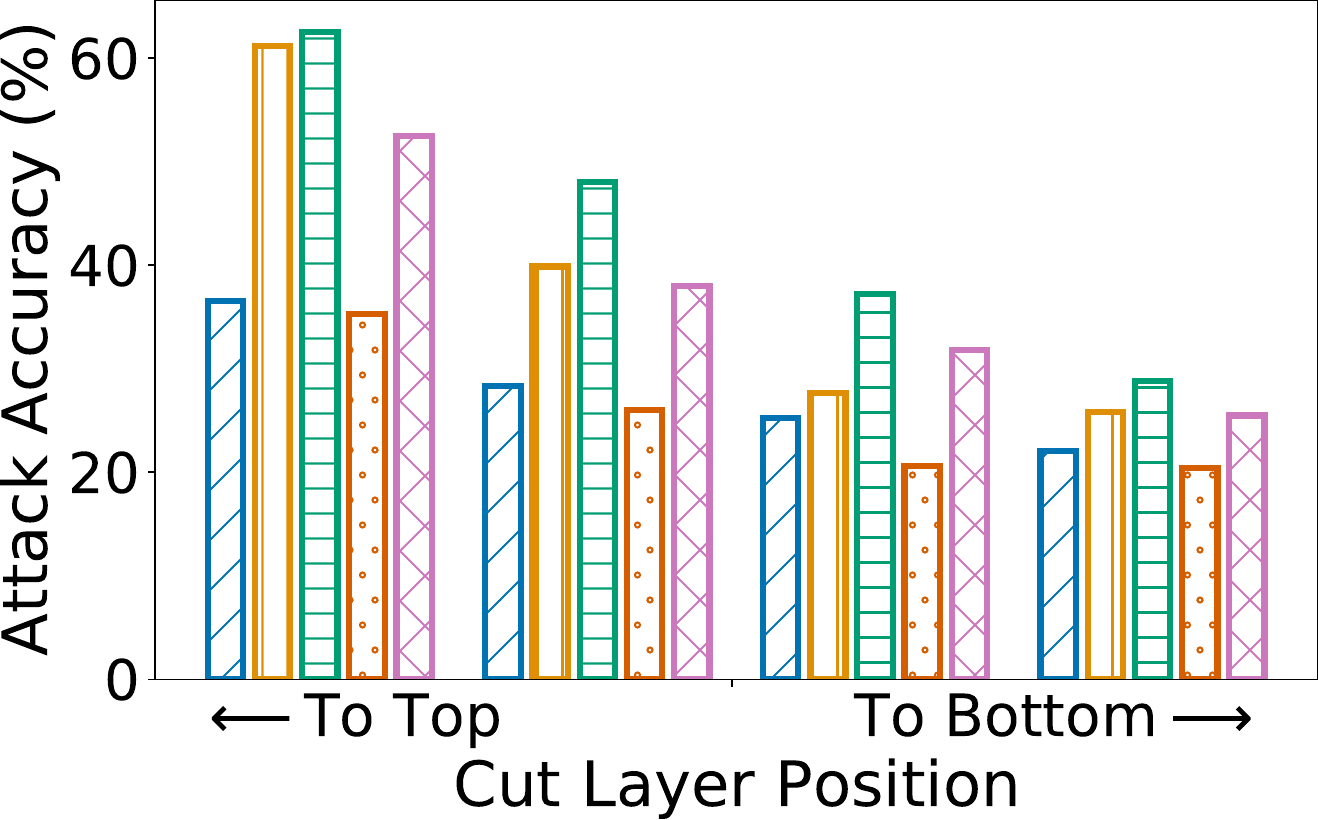}}
    \hspace{1mm}
    \subfloat[Completion, SVHN]{\includegraphics[width=0.24\textwidth]{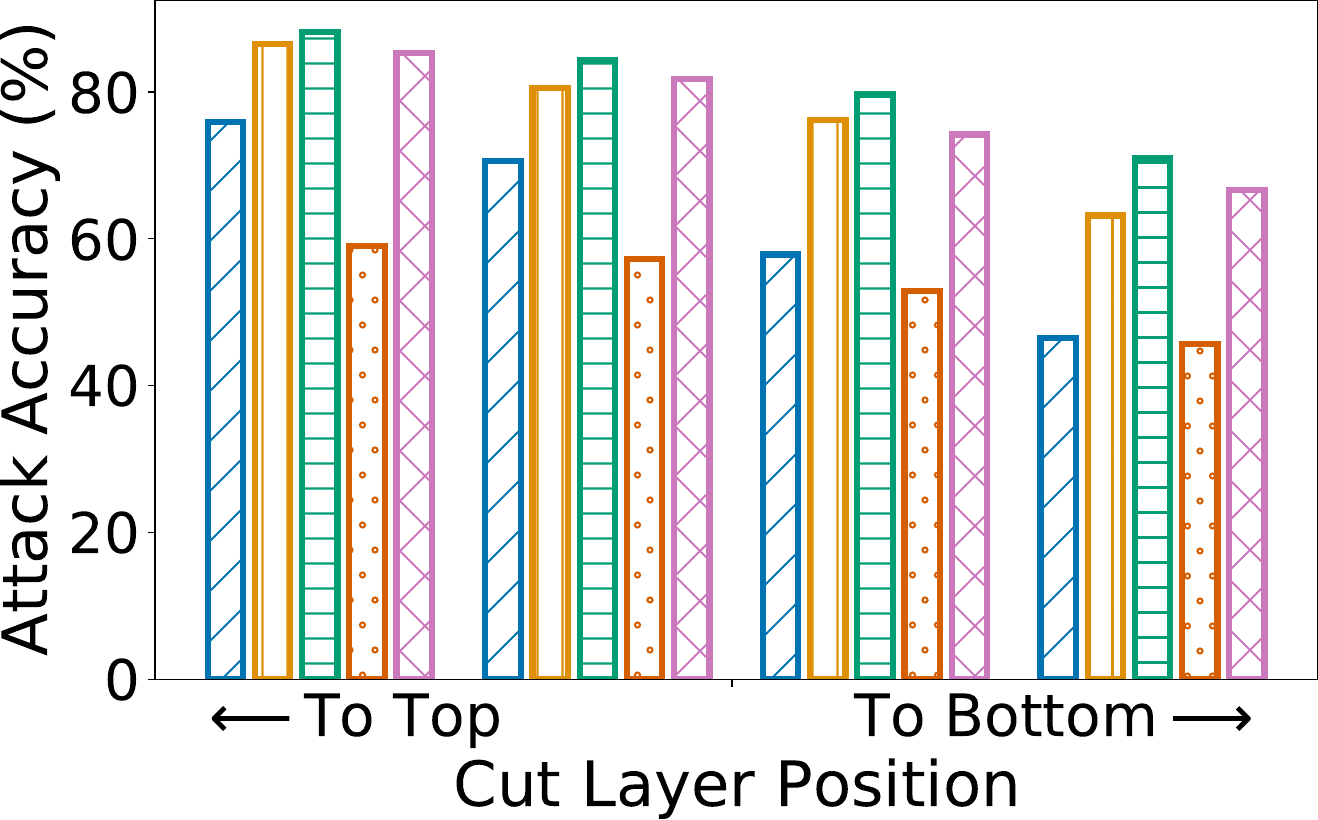}}
    \hspace{1mm}
    \subfloat[Completion, GTSRB]{\includegraphics[width=0.24\textwidth]{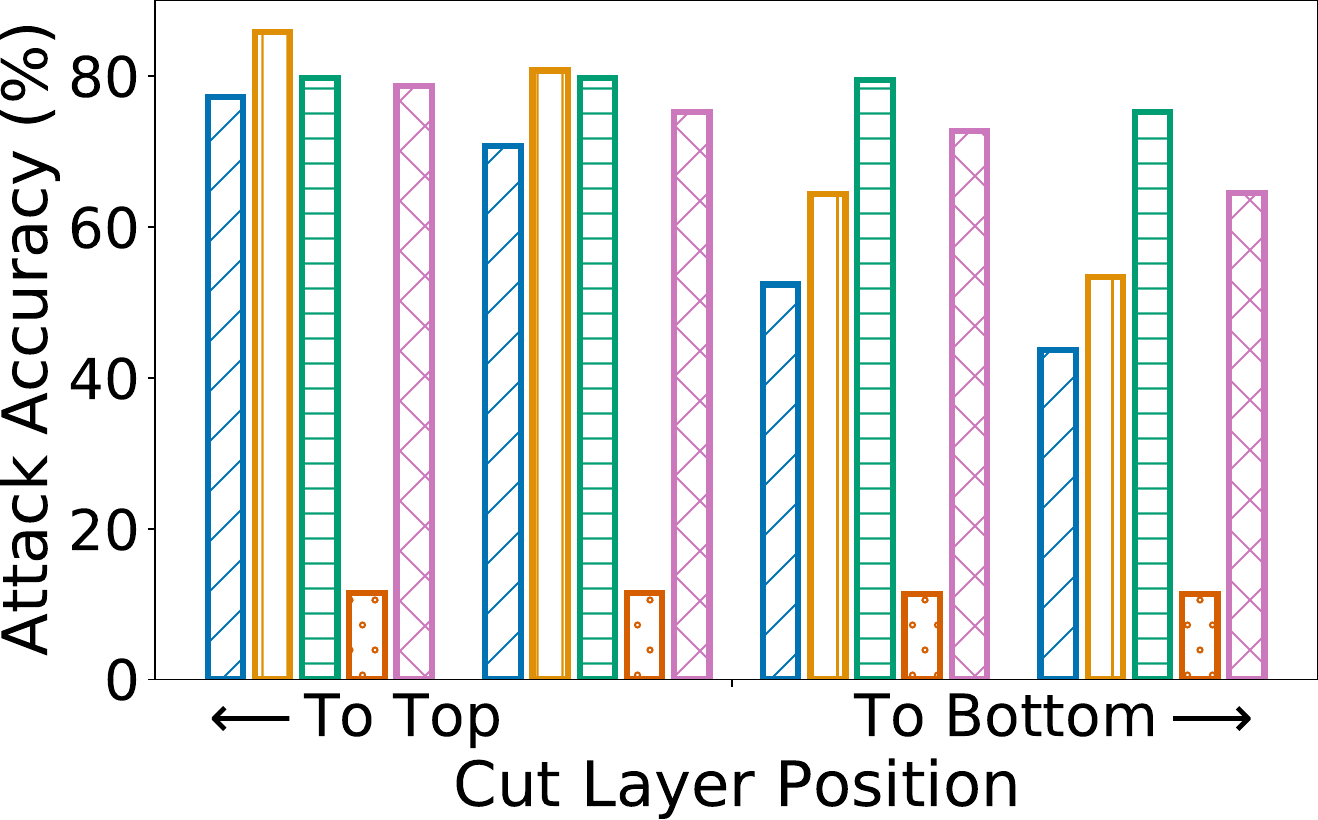}}
    \caption{Moving the cut layer closer to bottom models can effectively enhance other defenses against LIAs.}
    \label{fig:defense_strengthen}
\end{figure*}

\parahead{Confusional Autoencoder~(CAE).} Unlike the general defenses above, CAE~\cite{zou2024defending} is specifically designed to counter LIAs in VFL. It works by training an additional fake label generator to produce soft fake labels that differ from the original ones, allowing the VFL model to train on these fake labels. Since these fake labels are difficult to distinguish in distribution, they undermine the bottom model's ability to learn distinct labels, enabling CAE to effectively mislead LIAs that rely on the embedding-label assumption.

\parahead{Random Label Extension~(RLE).} RLE~\cite{qiu2023defending} is another specific defense against LIAs. It increases the dimensionality of labels and adds random noise, which extends the label space and increases uncertainty. This makes it more challenging for LIAs to accurately infer labels based on embeddings.

\parahead{}While applying the five defense strategies mentioned above, we simultaneously move the cut layer closer to the bottom model. The experimental results are shown in Fig.~\ref{fig:defense_strengthen}, with additional results for MNIST provided in Fig.~1 of Appendix~B. The figures show that advancing the cut layer consistently improves the effectiveness of all five defense strategies. Specifically, as the cut layer moves closer to the bottom model, the attack accuracy of both LIAs drops significantly under each defense. This demonstrates that advancing the cut layer not only directly mitigates LIA threats but also synergistically enhances the performance of existing defense strategies. We attribute this improvement primarily to the reduced label representation capability of the bottom model caused by advancing the cut layer. This effectively undermines the foundation of LIAs based on the embedding-label assumption, thereby boosting the effectiveness of other defenses.

\subsection{Robustness across Model Architectures}

The robustness of the defense effect achieved by moving the cut layer forward is also within our consideration. To this end, we evaluate our approach using three distinct model architectures: MLP, convolutional neural networks, and ResNet. The experiments evaluate performance from three perspectives, including networks with varying widths, networks of different depths, and the performance of different model architectures on the same dataset.

\begin{figure}[t!]
    \centering
    \includegraphics[width=0.9\columnwidth]{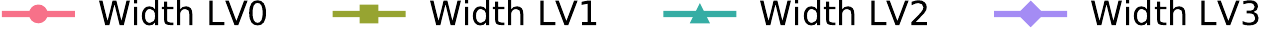}\\
    \subfloat[Cluster, MNIST, MLP]{\includegraphics[width=0.48\columnwidth]{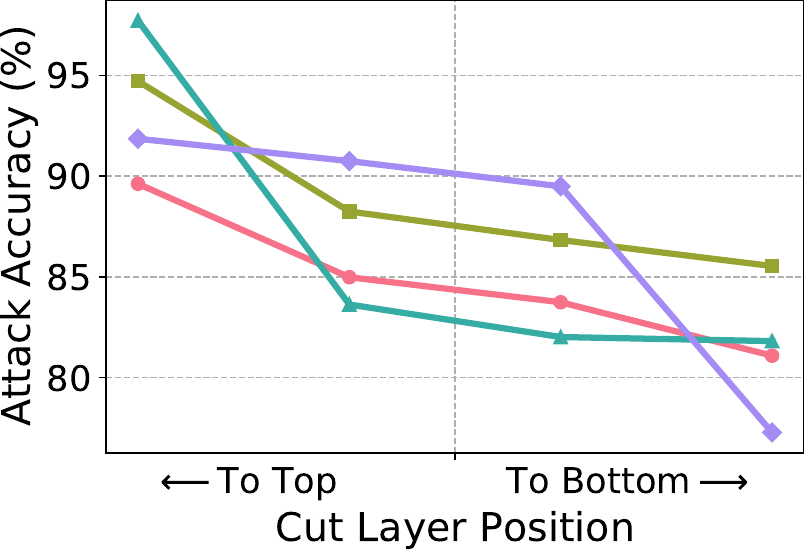}}
    \hspace{1mm}
    \subfloat[Completion, MNIST, MLP]{\includegraphics[width=0.48\columnwidth]{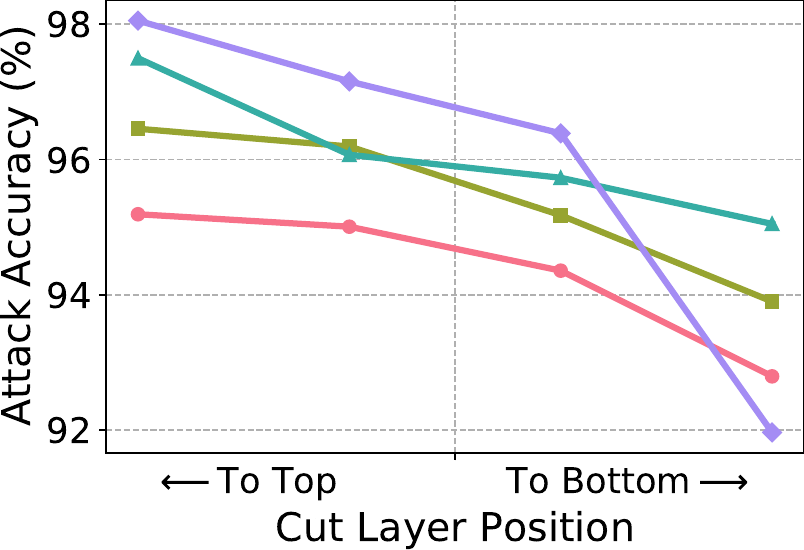}}\\
    \subfloat[Cluster, SVHN, CNN]{\includegraphics[width=0.48\columnwidth]{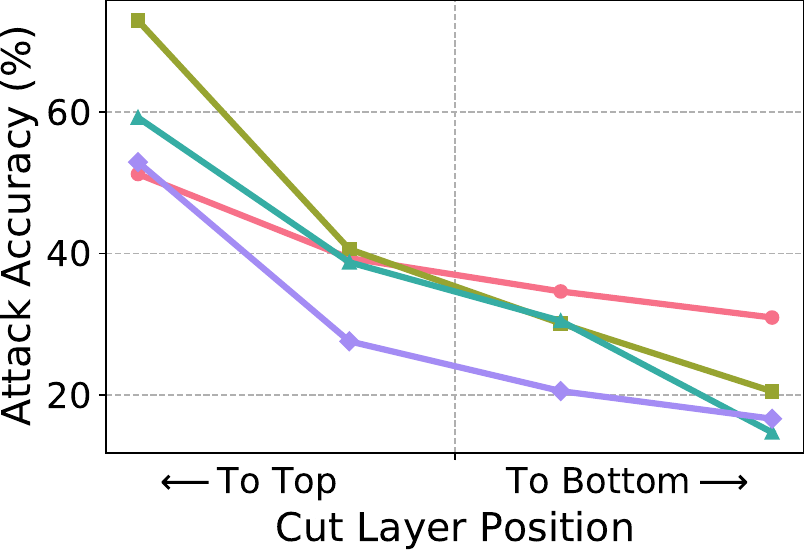}}
    \hspace{1mm}
    \subfloat[Completion, SVHN, CNN]{\includegraphics[width=0.48\columnwidth]{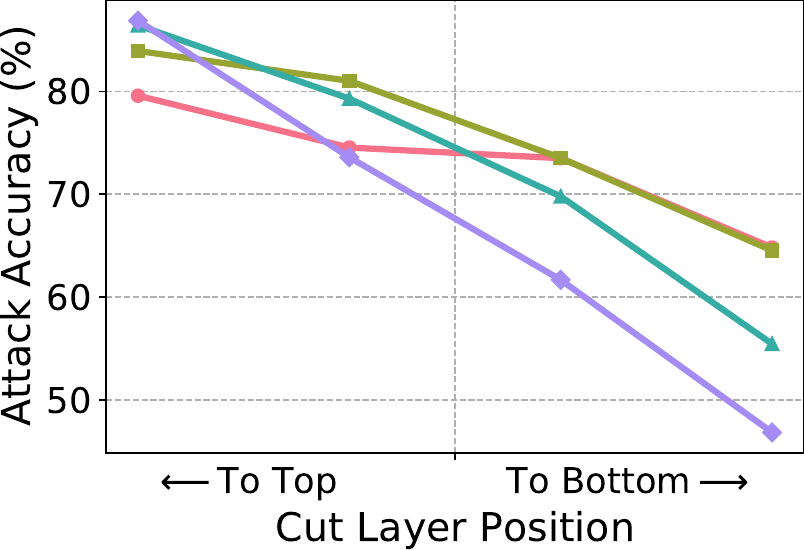}}\\
    \subfloat[Cluster, CINIC-10, ResNet]{\includegraphics[width=0.48\columnwidth]{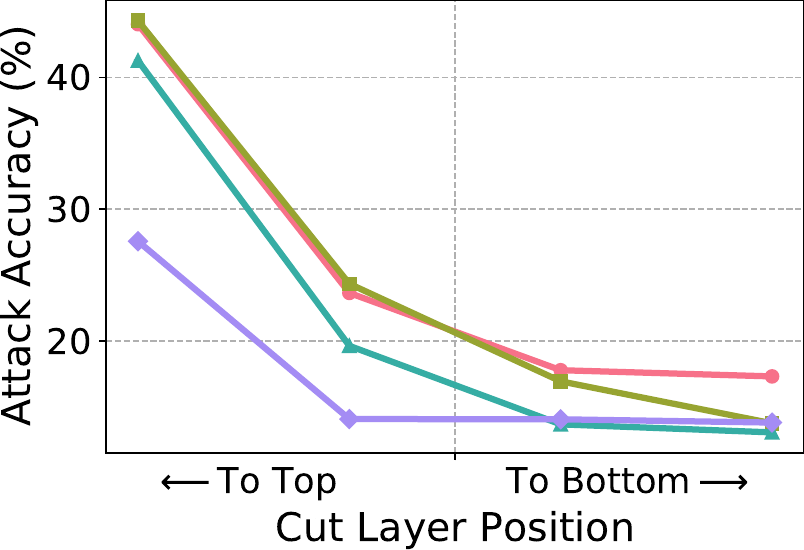}}
    \hspace{1mm}
    \subfloat[Completion, CINIC-10, ResNet]{\includegraphics[width=0.48\columnwidth]{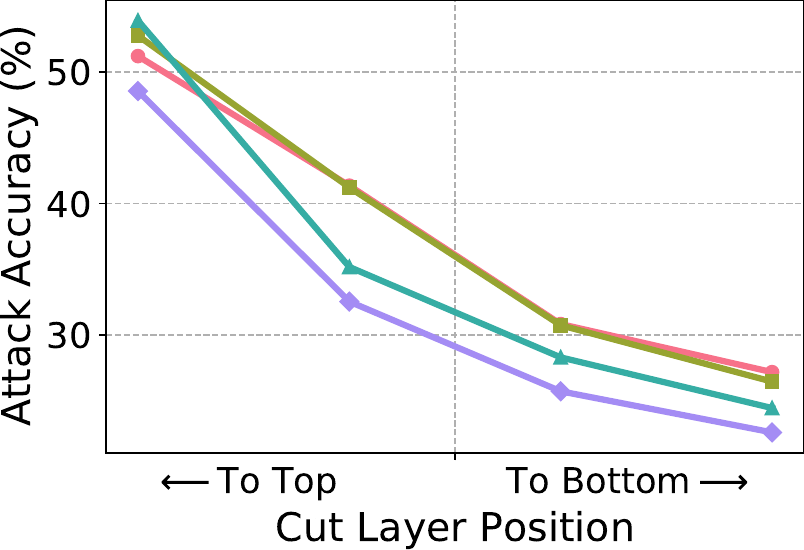}}
    \caption{Robustness of the defense effect across networks with varying widths.}
    \label{fig:fatten}
\end{figure}

\parahead{Varying Widths.} We first assess the defense effect of moving the cut layer on networks with different widths. Specifically, we adjust the width of each layer in the bottom and top models by multiplying it with a width factor ranging from 0.5 to 4.0. The experimental results for three representative datasets are presented in Fig.~\ref{fig:fatten}. It can be observed that, across all three datasets and model architectures, moving the cut layer closer to the bottom model consistently reduces the attack accuracy of both LIAs. Furthermore, although networks with different widths show varying degrees of accuracy decline when the cut layer is moved forward, the overall downward trends and proportions remain highly consistent across most datasets and model architectures. These results indicate that our proposed defense technique is robust across networks of varying widths.

\begin{figure}[t!]
    \centering
    \includegraphics[width=0.9\columnwidth]{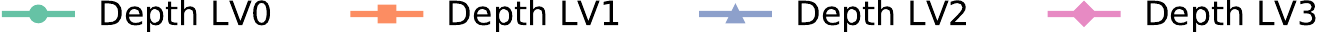}\\
    \subfloat[Cluster, MNIST, MLP]{\includegraphics[width=0.48\columnwidth]{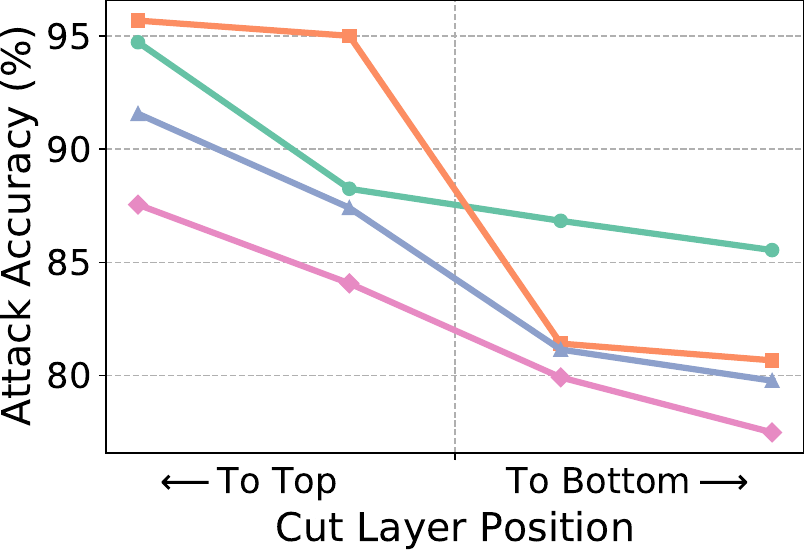}}
    \hspace{1mm}
    \subfloat[Completion, MNIST, MLP]{\includegraphics[width=0.48\columnwidth]{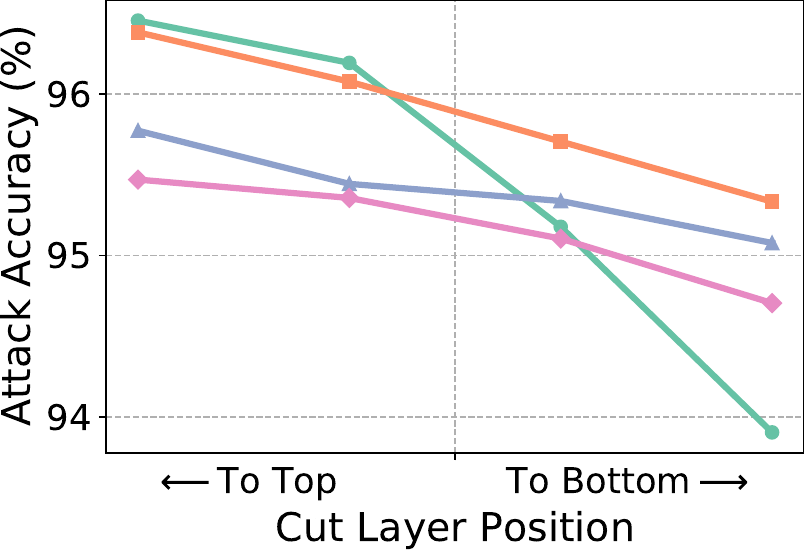}}\\
    \subfloat[Cluster, SVHN, CNN]{\includegraphics[width=0.48\columnwidth]{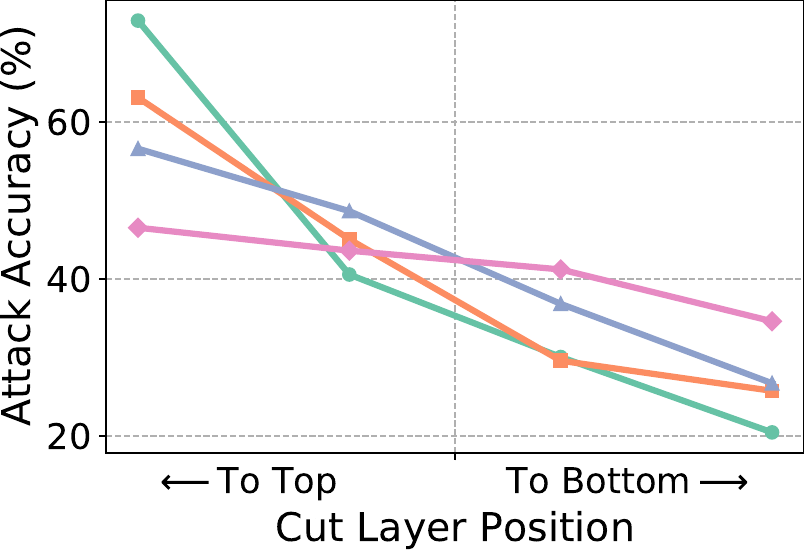}}
    \hspace{1mm}
    \subfloat[Completion, SVHN, CNN]{\includegraphics[width=0.48\columnwidth]{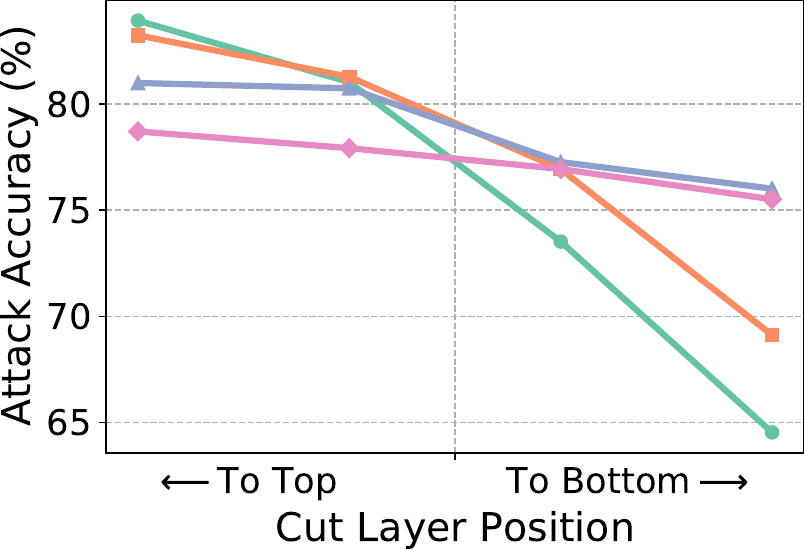}}\\
    \subfloat[Cluster, CINIC-10, ResNet]{\includegraphics[width=0.48\columnwidth]{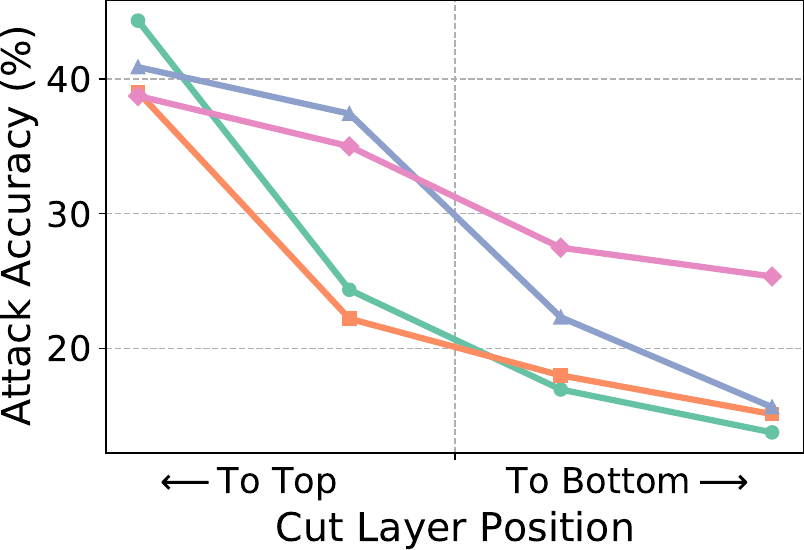}}
    \hspace{1mm}
    \subfloat[Completion, CINIC-10, ResNet]{\includegraphics[width=0.48\columnwidth]{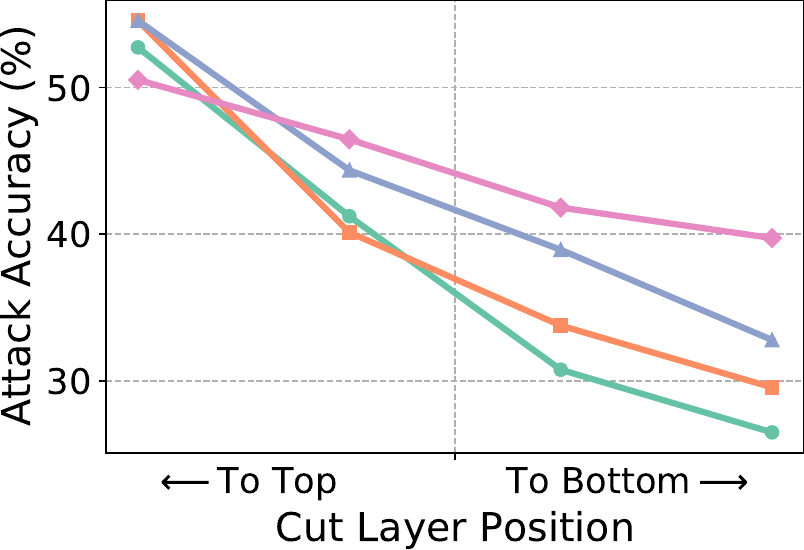}}
    \caption{Robustness of the defense effect across networks of different depths.}
    \label{fig:deepen}
\end{figure}

\parahead{Different Depths.} Next, we evaluate the defense effect of moving the cut layer on networks of different depths. We adjust the depth of the bottom and top models by adding or removing layers, varying the total number of layers from 10 to 18. The experimental results are presented in Fig.~\ref{fig:deepen}. We observe that, across all three datasets and model architectures, moving the cut layer forward consistently reduces the effectiveness of both types of LIAs. Similar to the results for networks with varying widths, we observe that models of different depths exhibit nearly identical trends in defensive effectiveness against LIAs when the cut layer is moved forward. These indicate that our proposed defense technique is robust across networks of varying depths.

\begin{figure}[t!]
    \centering
    \includegraphics[width=0.45\columnwidth]{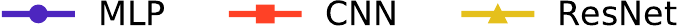}\\
    \subfloat[Cluster]{\includegraphics[width=0.48\columnwidth]{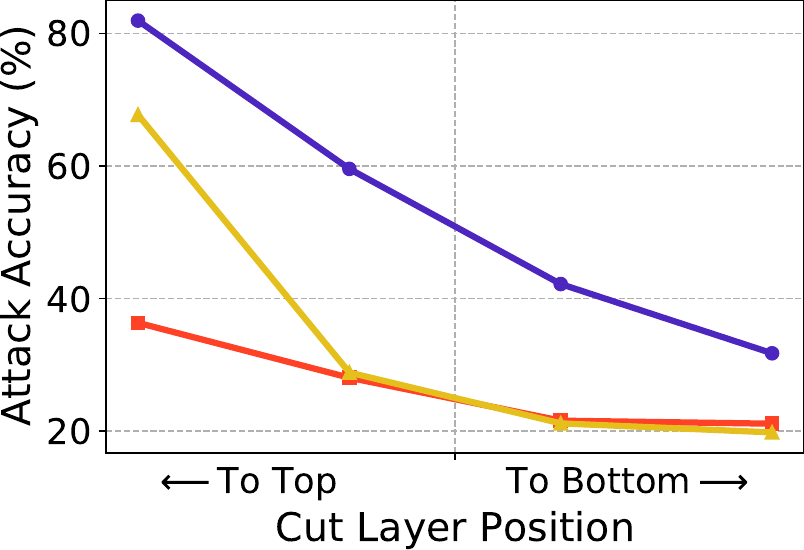}}
    \hspace{1mm}
    \subfloat[Completion]{\includegraphics[width=0.48\columnwidth]{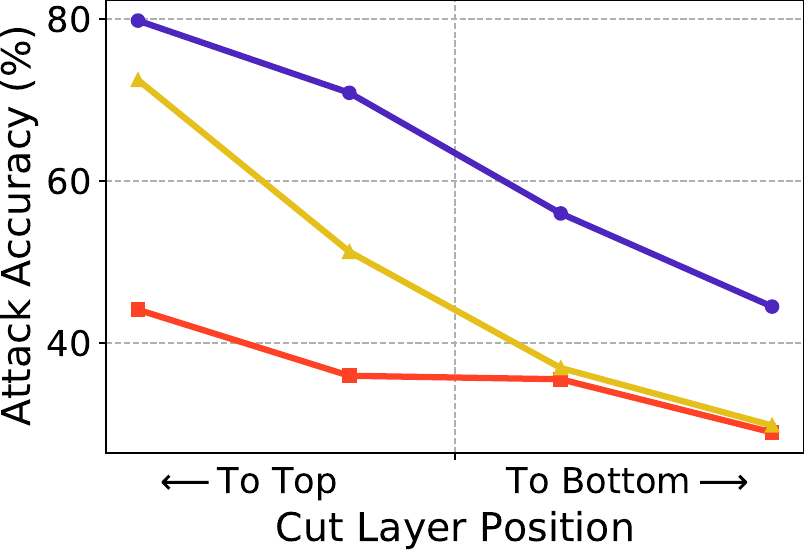}}
    \caption{Robustness of the defense effect across distinct model architectures.}
    \label{fig:different_model}
\end{figure}

\parahead{Distinct Architectures.} Finally, we investigate the performance of our defense technique across different model architectures. We conduct experiments on the CIFAR-10 dataset, with results shown in Fig.~\ref{fig:different_model}. The results demonstrate that the defense effectiveness achieved by moving forward the cut layer is not limited to a specific model or dataset. Instead, when the same dataset is trained using different model architectures, the defense technique remains effective, further validating its robustness.

\subsection{Discussion}

We do not intend to position ``advancing the cut layer to increase the proportion of the top model for defense'' as the defense strategy. Rather, we view it as a technique that can independently provide defense while also enhancing the effectiveness of other defense strategies. This distinction arises because most existing defenses do not directly modify the model structure; instead, they typically achieve privacy preservation by introducing additional perturbations or noise. In contrast, our approach is rooted in information-theoretic principles and achieves defense through simple, minimal adjustments to the model architecture. This technique is not only remarkably lightweight, but it also improves model predictive performance while providing robust privacy preservation. Moreover, due to its unique mechanism, we particularly highlight its ability to synergistically strengthen the effectiveness of other defense strategies.

In addition, we did not directly compare the effectiveness of the cut layer forward movement with other defense strategies in the preceding discussion. This is because we believe that advancing the cut layer inevitably alters the model's structure, whereas most existing defense strategies typically do not modify the model architecture. Such a comparison could introduce bias by failing to control for this variable. Nevertheless, the experimental results in Table~\ref{tab:defense_move} and Fig.~\ref{fig:defense_strengthen} show that simply advancing the cut layer by one layer achieves defense performance comparable to other defense strategies, and even outperforms the best defenses on some datasets. When the cut layer is advanced by three layers, its defense performance significantly exceeds that of other strategies. Moreover, this defense technique can be regarded as \textbf{zero-overhead}, as it does not introduce additional models or perturbations but simply adjusts the proportion between the bottom and top models, thereby incurring no extra computational overhead. These findings indicate that advancing the cut layer is a highly effective defense technique against LIAs.

Whether continuously advancing the cut layer is feasible is also a matter worthy of discussion. We believe that continuously advancing the cut layer is not the optimal solution for privacy preservation in VFL. This is because VFL faces not only the risk of LIAs but also threats from other types of attacks. For example, in addition to the risk of label leakage for the active party, passive parties are also exposed to feature leakage risks. Specifically, the active party may infer features from the embeddings uploaded by passive parties, enabling feature inference attacks~\cite{qiu2024hashvfl,vu2023active,fu2025privacy}. Therefore, we argue that excessively advancing the cut layer may increase the risk of feature inference attacks. This is because moving the cut layer closer to the bottom model results in fewer layers for the bottom model to process sample features, leading to embeddings that retain more raw feature information. Consequently, we recommend a balanced approach when adjusting the cut layer, and we will investigate its trade-offs in our future work.


\section{Conclusion}

In the paper, we have critically examined the foundational embedding-label assumption underlying existing LIAs in VFL. By leveraging mutual information, we have presented the first observation of the model compensation in VFL and clarified the distinct roles of bottom and top models. We have also theoretically proved that mutual information between model layer outputs and labels increases with layer depth. Building on this insight, we have conducted task reassignment experiments to demonstrate that the embedding-label assumption and existing LIAs are vulnerable. This vulnerability arises because their high attack accuracy largely results from the illusion of a strong natural alignment between sample features and labels. When this alignment is disrupted, LIA performance degrades sharply, potentially even falling below random guessing. Furthermore, we propose a concise and effective defense technique: moving cut layers forward to weaken bottom models. This approach not only provides robust defense against LIAs but also improves the overall predictive performance of VFL models, synergistically enhancing the effectiveness of existing defense strategies. We believe our work offers valuable insights for understanding and defending against LIAs in VFL.

\section*{Acknowledgments}
This work was supported by the National Natural Science Foundation of China under Grant 62572007.

\small
\bibliographystyle{IEEEtranN}
\bibliography{references}

\end{document}